\documentclass[letterpaper, 10 pt, conference]{ieeeconf}  

\IEEEoverridecommandlockouts                              

\usepackage[utf8]{inputenc} 
\usepackage[T1]{fontenc}    
\usepackage{hyperref}       
\usepackage{url}            
\usepackage{booktabs}       
\usepackage{amsfonts}       
\usepackage{nicefrac}       
\usepackage{microtype}      
\usepackage{xcolor}         
\usepackage{amsmath}
\usepackage{tikz} 
\usepackage{ntheorem}
\usetikzlibrary{external}
\usetikzlibrary{positioning}
\usepackage{array}
\usepackage{textcomp}
\usepackage{pgfplots}

\DeclareMathOperator*{\argmin}{arg\,min}
\pgfplotsset{compat=1.16}
\usepackage[english]{babel}
\usepackage{graphicx}
\usepackage{wrapfig}
\usepackage[font=small]{caption}
\usepackage{subcaption}
\usepackage{sidecap}
\usepackage{graphicx,caption}
\let\labelindent\relax
\usepackage[shortlabels]{enumitem} 
\setlist[enumerate,1]{leftmargin=0.5cm}
\usepackage{flushend}
\usepackage{cite}
\usepackage{algorithm}
\usepackage{algpseudocode}

\title{\LARGE \bf
An Optimal Transport Approach for Network Regression
}

\author{Alex G. Zalles, Kai M. Hung, Ann E. Finneran, Lydia Beaudrot, and C\'esar A. Uribe
\thanks{*This work is supported by the National Science Foundation No. 2213568.}
\thanks{AGZ is with the Department of Computational Applied Mathematics and Operations Research,
        Rice University, Houston, TX 77005, USA. {\tt\small agz2@rice.edu}. KMH is with the Department of Computer Science, Rice University, Houston, TX 77005, USA. {\tt\small kai.hung@rice.edu}. AEF and LB are with the Department of Biosciences, Rice University, Houston, TX 77005, USA.
        {\tt\small \{annie.finneran, beaudrot\}@rice.edu}. CAU is with the Department of Electrical and Computer Engineering, Rice University, Houston, TX 77005, USA.
        {\tt\small cauribe@rice.edu}.}%
}

\begin{document}

\maketitle
\thispagestyle{empty}
\pagestyle{empty}

\begin{abstract}

We study the problem of network regression, where one is interested in how the topology of a network changes as a function of Euclidean covariates. We build upon recent developments in generalized regression models on metric spaces based on Fr\'echet means and propose a network regression method using the Wasserstein metric. We show that when representing graphs as multivariate Gaussian distributions, the network regression problem requires the computation of a Riemannian center of mass (i.e., Fr\'echet means). Fr\'echet means with non-negative weights translates into a barycenter problem and can be efficiently computed using fixed point iterations. Although the convergence guarantees of fixed-point iterations for the computation of Wasserstein affine averages remain an open problem, we provide evidence of convergence in a large number of synthetic and real-data scenarios.  
Extensive numerical results show that the proposed approach improves existing procedures by accurately accounting for graph size, topology, and sparsity in synthetic experiments. Additionally, real-world experiments using the proposed approach result in higher Coefficient of Determination ($\bf{R^{2}}$) values and lower mean squared prediction error (MSPE), cementing improved prediction capabilities in practice. 

\end{abstract}

\newtheorem{theorem}{Theorem}
\newtheorem{proposition}{Proposition}
\newtheorem{remark}{Remark}

\section{Introduction}

At the core of many classical~\cite{kopp1963linear} and modern data-based control systems, one usually finds some form of regression analysis~\cite{baggio2021data}. Nevertheless, regression is typically studied in Euclidean spaces, where regressors and outputs are real multivariate values~\cite{linearRegression}. For example, in linear systems identification, relationships between predictor variables and their outputs are quantified by minimizing least squares error. 
As data modalities within modern control systems grow, extensions to non-Euclidean regression models are needed, e.g., estimation, inference, learning, and control for graph data and Euclidean covariates~\cite{chami2022machine, otherGraphPrediction}. Successful applications range from brain imaging~\cite{brainConnectivity} where neurons are clustered, and we model their inter-connectivity as an output of age, ecology~\cite{ecologyApplication}, scheduling~\cite{tripPrediction}, estimation~\cite{nedic2017fast}, to control~\cite{bemporad2010networked}. 


In graph prediction for networked control systems, prior work has successfully developed regression models using the Frobenius Norm metric for graphs~\cite{mainPaper}, implementing models to quantify inter-state traffic as a response to changing COVID-19 case numbers. 
However, recent results suggest that  Wasserstein distances better encapsulate a graph structure for data-based learning and control systems~\cite{petric2019got}.  By representing graphs as multivariate Gaussian distributions~\cite{dong2016learning}, we can define a distance between graphs using the Wasserstein distance between their respective Gaussian representations. Evidence suggests that Wasserstein distances allow for the prioritization of global connectivity, which contributes to a more robust metric~\cite{maretic2021fgot}. These claims are supported by recent works studying graph averages and interpolation through Wasserstein barycenters~\cite{haasler2023bureswasserstein} and employing learning methods in Gromov-Wasserstein computations~\cite{brogatmotte2022learning}. 

Recent research has enabled and developed the formal definition of regression models in general metric spaces using Fr\'echet means. Fr\'echet means, and variances extend notions of averages and standard deviations to metric spaces, which allows for the definition of regression models beyond Euclidean spaces~\cite{chen2022uniform, Frechet1948, petersen2019frechet}. Regression in Wasserstein space has been recently studied using quantile functions and empirical one-dimensional measures~\cite{fan2021conditional, zhou2023wasserstein}. 
Later,~\cite{fan2021conditional} bridged theoretical formulations with computational approaches for discretizing higher dimensional distributions, where statistical consistency was shown.


In this paper, we develop a network regression model where we leverage two main ideas: 1) Fr\'echet means for regression in Wasserstein spaces, and 2) Network (graph) representation as multidimensional Gaussian distributions. By combining the effective weight function as derived in Fr\'echet regression models~\cite{mainPaper, petersen2019frechet} and the Wasserstein metric~\cite{petric2019got, maretic2021fgot} for graph comparison, we demonstrate better performances of Wasserstein-based network regression models when compared with traditional Euclidean-based methods. 
Errors in the proposed method scale better with respect to the number of nodes compared with state-of-the-art methods.
Moreover, the Wasserstein-based method results in smaller prediction errors and greater model fitness in the $R^{2}$ coefficient.  We show the regressor's improved predictions in graph swelling, graph interpolation, and utilization of a graph Laplacian's spectral properties. These synthetic experiments are then extended to real-world applications on taxi-cab data in Manhattan as a response to COVID-19 cases, where the Wasserstein regressor demonstrates improved model fitness and lower prediction error while maintaining efficiency. 

This paper is organized as follows. Section \ref{Section:2} outlines the global regressor for general metric spaces and, in particular, network spaces, defining both the Wasserstein and Frobenius distances over these networks and providing a simple encouraging example. Section \ref{Section 3} then provides computational frameworks for solving the Wasserstein regressor. Section \ref{sec:numerics} shows different experiments for the proposed Wasserstein-based regression and compares them with classical Frobenius-based approaches. Finally, Section \ref{sec: conclusion} outlines further growth points for our methods to increase their applicability, and the Appendix includes additional experiments based on those in the main sections.






\section{Regression on Network Metric Spaces} \label{Section:2}


\subsection{Fr\'echet averages and Regression}

Consider a random pair $(X,G) \sim F$, where $F$ is a joint distribution, $X$ takes value in $\mathbb{R}^p$ and $G=(V,E,W)$ is a random graph with a fixed node set $V$, edge set $E\subseteq V \times V$, and $W \in \mathbb{R}^{|V|\times |V|}_{\geq0}$ as the set of bounded non-negative edge weights, which contains the randomness of the system. Moreover, we assume $G$ takes value in a metric space $(\mathcal{G},d)$, where $\mathcal{G}$ is the space of graphs with $|V|$ nodes, metriced by $d$, in our case the Frobenius and Wasserstein distances, with corresponding marginal distributions $F_{X}$ and $F_{G}$, for which the conditional distributions $F_{X|G}$ and $F_{G|X}$ exist.  Extending traditional concepts of mean and variances to metric spaces, the Fr\'echet mean and variance~\cite{Frechet1948} are defined as
\[
G_{\oplus} = \argmin_{G \in \mathcal{G}} \mathbb{E}[d^2(X, G)], \ \text{and} \ \; \; V_{\oplus} = \mathbb{E}[d^2(X, G_{\oplus})].
\]

In the Euclidean setting, for jointly distributed random variables $X$ and $Y$, the conditional distribution is $\mathbb{E}[Y \mid X= x] = \argmin_{y\in \mathbb{R}} \mathbb{E}[(Y-y)^2 \mid X= x]$. The authors in~\cite{mainPaper,petersen2019frechet} propose a conditional Fr\'echet mean as a natural extension to network-valued and other metric space-valued responses, where $(Y-y)^2$ is replaced by $d^2(G,w)$ for some $w$ in the metric space. Thus, the corresponding regression function of $G$ given $X = x$ is defined as 
\[
    m(x) := \argmin_{w \in \mathcal{G}} \mathbb{E}[d^{2}(G, w) | X = x].
\]
Moreover, by characterizing the regression function as a weighted least square problem, the authors in~\cite{mainPaper} propose a global Fr\'echet Regression model given $X = x$ as the affine average
\[
    m_{G}(x) := \argmin_{w \in \mathcal{G}}  \mathbb{E}[s_{G}(X, x)d^{2}(G, w)],
\]
for a weight function $s_{G}(X, x) = 1 + (X - \mu)^{T} \Sigma^{-1}(x - \mu)$, which is formulated to replicate Euclidean regression properties in metric spaces~\cite[Section 2.2]{petersen2019frechet}. Here, $\mu=\mathbb{E}[X]$ and $\Sigma =\text{cov}(X)$. Similarly, when a finite set of i.i.d. pairs $(X_i,G_i)\sim F$ for $i=1,\cdots,n$ is available, the model becomes the empirical regressor
\begin{align}\label{eq:emp_reg}
    \hat{m}_{G}(x) := \argmin_{w \in \mathcal{G}}  \frac{1}{n}\sum_{i = 1}^{n}s_{iG}(X_{i}, x)d^{2}(G_{i}, w),
\end{align}
where the sample weight function is defined as $s_{iG}(X_i,x) =  1 + (X_{i} - \bar{X})^{T}\hat{\Sigma}^{-1}(x - \bar{X})$, $\bar{X} = n^{-1} \sum_{i = 1}^{n}X_{i}$ is the sample mean, and $\hat{\Sigma} = n^{-1}\sum_{i = 1}^{n}(X_{i} - \bar{X})(X_{i} - \bar{X})^{T}$ is the sample covariance matrix. Through incorporating a smoothing kernel~\cite{Rasmussen2004}, we define the Local Regression model~\cite{mainPaper} to reduce bias from sampling effects on our data distribution, outlined in Appendix~\ref{app:A}.

Next, we describe the two metrics $d$ with which we will study the network regression problem.

\subsection{Metrics for Regression on Networks}

The space of networks can be defined for many metrics, each encapsulating network difference separately~\cite{graphMeasures}. We will focus on the Frobenius and the Wasserstein metrics and demonstrate their formulations for simple, undirected graphs with real-valued edge weights. The Frobenius Norm is a baseline metric for initial network regression models~\cite{mainPaper}. However, recent results suggest that Wasserstein distances outperform Frobenius distances when comparing network structures. By representing graphs as signals, the Wasserstein distance can better capture global graph structure by measuring the discrepancy in lower graph frequencies~\cite{petric2019got}. Further exploration into graph signal processing can be found in~\cite{dong2016learning, Shuman_2013}. Therefore, one of our contributions is to propose using Wasserstein metrics in network regression problems and to present a set of algorithms and their computational considerations, showing empirical evidence of the improved performance of Wasserstein metrics versus Frobenius metrics.

\underline{\textit{Frobenius Norm~\cite{mainPaper}:}} All labeled  graphs with no self-loops, no multi-edges, and non-negative edge weights $w_{ij}$ have a one-to-one correspondence with their Laplacians $L= (L_{ij})$
\[
L_{ij} =
\left\{
    \begin{array}{lr}
        -w_{ij}, & \text{ if } i \neq j \\
        \sum_{k \neq i} w_{ik}, & \text{if } i = j,
    \end{array}
\right.
\]
which are always positive semi-definite matrices by definition~\cite{andreotti2018multiplicity}. Given two graphs $G_1$ and $G_2$ with their corresponding graph Laplacians, $L_1$ and $L_2$, the power Frobenius Norm, $d_{F,\alpha}(G_1,G_2)$, between graphs $G_1$ and $G_2$ is defined as
\begin{align*}
    d_{F,\alpha}(G_{1}, G_{2}) &{=} d_{F}(F_\alpha(L_{1}), F_\alpha(L_{2})){=} \|F_\alpha(L_{1}) {-} F_\alpha(L_{2})\|_{F} ,
\end{align*}
and $F_\alpha(S) = U\Lambda^\alpha U^T$, where $U$ is an orthogonal matrix, and $\Lambda$ is a diagonal matrix, representing the eigendecomposition of $S$, and $\alpha>0$, a power applied to the eigenvalues contained in $\Lambda$ for metric scaling. Setting $\alpha=1$, one recovers the Frobenius metric~\cite{petersen2019frechet}.

Recently, the authors in~\cite{petric2019got} showed several advantages of using Wasserstein distances with respect to classical Frobenius distances for capturing the geometric properties of graphs. So, we explicitly define the Wasserstein distance between two graphs, which builds on interpretations of graphs as elements of multidimensional distributions of signals as proposed in~\cite{picka2006gaussian}. 

\underline{\textit{Wasserstein Distance~\cite{petric2019got}:}} The $2$-Wasserstein distance between graphs $G_{1}, G_{2}$ is defined as 
    \begin{align*}
        d_{W}(G_{1}, G_{2}) &= W_{2}^{2}(\nu^{G_{1}}, \nu^{G_{2}}) \\ 
        &= \inf_{T_{\#}\nu^{G_{1}} = \nu^{G_{2}}} \int_{\mathbb{R}^{|V|}} \|x - T(x)\|^{2}d\nu^{G_{1}},
    \end{align*}
    
\noindent where $\nu^{G_{i}} = \mathcal{N}(0, L_{i}^{\dagger})$ with $L_{i}^{\dagger}$ denoting the pseudo-inverse of Laplacian $L_{i}$~\cite{pseudoInverse}. Given that $\nu^{G_{1}}$ and $\nu^{G_{2}}$ are zero-mean Gaussian distributions, the authors in~\cite{wassClosedForm} showed their $2$-Wasserstein Distance has the closed-form
\[
    W_{2}^{2}(\nu^{G_{1}}, \nu^{G_{2}}) = Tr(L_{1}^{\dagger} + L_{2}^{\dagger}) -2Tr(\sqrt{L_{1}^{\dagger/2}L_{2}^{\dagger}L_{1}^{\dagger/2}}).
\]
\begin{remark}
Note that the $2$-Wasserstein Distance between graphs reduces to the Bures-Wasserstein distance between positive (semi)definite matrices \cite{otherGraphPrediction,zheng2023barycenter}.
\end{remark}

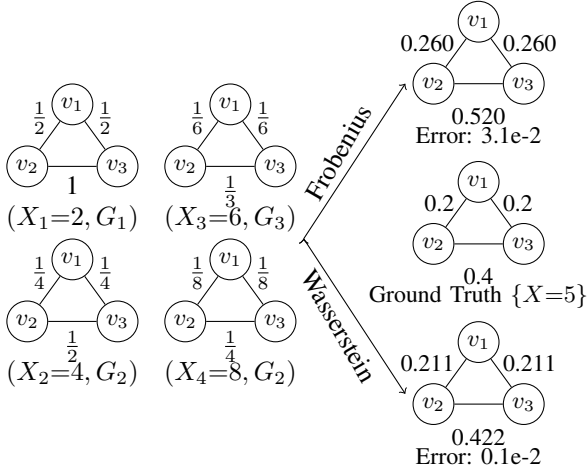
\begin{figure}
\centering
\begin{tikzpicture}[xscale = 0.3, yscale = 0.55]
\begin{scope}[every node/.style={circle, inner sep=2pt}, minimum size = 0.1em]
\node[draw] (a1) at (0, 2.5) {\small $v_{1}$};
\node[draw] (a2) at (-2, 1.0) {\small $v_{2}$};
\node[draw] (a3) at (2, 1.0) {\small $v_{3}$};
\node (a4) at (0, -0.25) {$(X_1{=}2,G_{1})$};

\node[draw] (b1) at (0, -1.25) {\small $v_{1}$};
\node[draw] (b2) at (-2, -2.75) {\small $v_{2}$};
\node[draw] (b3) at (2, -2.75) {\small $v_{3}$};
\node (b4) at (0, -4) {$(X_2{=}4,G_{2})$};

\node[draw] (c1) at (7, 2.5) {\small $v_{1}$};
\node[draw] (c2) at (5, 1.0) {\small $v_{2}$};
\node[draw] (c3) at (9, 1.0) {\small $v_{3}$};
\node (c4) at (7, -0.25) {$(X_3{=}6,G_{3})$};

\node[draw] (d1) at (7, -1.25) {\small $v_{1}$};
\node[draw] (d2) at (5, -2.75) {\small $v_{2}$};
\node[draw] (d3) at (9, -2.75) {\small $v_{3}$};
\node (d4) at (7, -4) {$(X_4{=}8,G_{2})$};
\end{scope}
\begin{scope}[every edge/.style={draw=black}]

\draw  (a1) edge node[above left, yshift = -1mm, xshift = 1mm] {$\frac{1}{2}$} (a2);
\draw  (a1) edge node[above right, yshift = -1mm, xshift = -1mm] {$\frac{1}{2}$} (a3);
\draw  (a2) edge node[below] {1} (a3);

\draw  (b1) edge node[above left, yshift = -1mm, xshift = 1mm] {$\frac{1}{4}$} (b2);
\draw  (b1) edge node[above right, yshift = -1mm, xshift = -1mm] {$\frac{1}{4}$} (b3);
\draw  (b2) edge node[below] {$\frac{1}{2}$} (b3);

\draw  (c1) edge node[above left, yshift = -1mm, xshift = 1mm] {$\frac{1}{6}$} (c2);
\draw  (c1) edge node[above right, yshift = -1mm, xshift = -1mm] {$\frac{1}{6}$} (c3);
\draw  (c2) edge node[below] {$\frac{1}{3}$} (c3);

\draw  (d1) edge node[above left, yshift = -1mm, xshift = 1mm] {$\frac{1}{8}$} (d2);
\draw  (d1) edge node[above right, yshift = -1mm, xshift = -1mm] {$\frac{1}{8}$} (d3);
\draw  (d2) edge node[below] {$\frac{1}{4}$} (d3);
\end{scope}

\begin{scope}[every node/.style={circle, inner sep=2pt}, minimum size = 0.1em]
\node[draw] (e1) at (18, 4.5) {\small $v_{1}$};
\node[draw] (e2) at (16, 3) {\small $v_{2}$};
\node[draw] (e3) at (20, 3) {\small $v_{3}$};
\node (e5) at (18, 1.75) {\small Error: 3.1e-2};

\end{scope}
\begin{scope}[every edge/.style={draw=black}]

\draw  (e1) edge node[above left, yshift = -1mm, xshift = 1mm] {\small 0.260} (e2);
\draw  (e1) edge node[above right, yshift = -1mm, xshift = -1mm] {\small 0.260} (e3);
\draw  (e2) edge node[below, yshift = -2mm] {\small 0.520} (e3);

\end{scope}
\begin{scope}[every node/.style={circle, inner sep=2pt}, minimum size = 0.1em]
\node[draw] (g1) at (18, -3.25) {\small $v_{1}$};
\node[draw] (g2) at (16, -4.75) {\small $v_{2}$};
\node[draw] (g3) at (20, -4.75) {\small $v_{3}$};
\node (g5) at (18, -6) {\small Error: 0.1e-2};

\end{scope}
\begin{scope}[every edge/.style={draw=black}]

\draw  (g1) edge node[above left, yshift = -1mm, xshift = 1mm] {\small 0.211} (g2);
\draw  (g1) edge node[above right, yshift = -1mm, xshift = -1mm] {\small 0.211} (g3);
\draw  (g2) edge node[below, yshift = -2mm] {\small 0.422} (g3);
\end{scope}

\begin{scope}[every node/.style={circle, inner sep=2pt}, minimum size = 0.1em]
\node[draw] (h1) at (18, 0.625) {\small $v_{1}$};
\node[draw] (h2) at (16, -0.875) {\small $v_{2}$};
\node[draw] (h3) at (20, -0.875) {\small $v_{3}$};
\node (h5) at (18, -2.125) {\small Ground Truth $\{X{=}5\}$};

\end{scope}
\begin{scope}[every edge/.style={draw=black}]

\draw  (h1) edge node[above left, yshift = -1mm, xshift = 1mm] {\small 0.2} (h2);
\draw  (h1) edge node[above right, yshift = -1mm, xshift = -1mm] {\small 0.2} (h3);
\draw  (h2) edge node[below, yshift = -2mm] {\small 0.4} (h3);
\end{scope}

\begin{scope}[every node/.style={sloped,anchor=south,auto=false}]
    \draw[->](10.25, -0.75) edge node[rotate = 20] {Frobenius} (14.75, 3);
    \draw[->](10.25, -0.75) edge node[below, rotate = -20] {Wasserstein} (14.75, -4.5);
\end{scope}
\end{tikzpicture}
\vspace{-0.6cm}
\caption{We train our global network regression models over $\{X_i,G_i\}_{i = 1}^4$ pairs where $G_i$ is the response and $X_i$ is the predictor. Then, we predict the graphs with predictor $x = 5$. The Frobenius regressor produces a graph (top) with thirty times the error than our Wasserstein regressor (bottom).}
\label{fig:1}
\vspace{-0.4cm}
\end{figure}

We introduce our numerical experiments by presenting a toy example for network regression on simple graphs initially developed in~\cite{mainPaper}. Figure~\ref{fig:1} shows the results of network regression where the Wasserstein-based regressor outperforms the Frobenius-based regressor. Specifically, we have four random pairs $\{X _i,G_i\}_{i=1}^4$ independently observed, with weights shown next to the corresponding edges. Each graph $G_i$ has an associated covariate $X_i$, which is accounted for in the weights of each sample graph during prediction. We seek to estimate $G$ at $x=5$ by finding the conditional expectation of $G$ with response to $x=5$ through the regression models defined above, where we find the sample graph weights are $s_{iG} = 0.25$ for $i=1,\ldots, 4$ and $s_{iL} = 0.5$ for $i=2, 3$, demonstrating importantly that the weights sum to one. The unknown ground truth model is $w_{1,2}=w_{1,3}=1/X$ and $w_{2,3}=2/X$, thus we expect $w_{1,2}=w_{1,3}=0.2$ and $w_{2,3}=0.4$. 

\section{Computational Aspects of Wasserstein Network Regressions}\label{Section 3}

The empirical regressor in~\eqref{eq:emp_reg} takes the form of an affine combination of convex functions where the weights are defined by the function $s_{kG}$ since $\mathbb{E}[s_{kG}(X,x)]=1$. Frobenius regression models thus require solving a convex quadratic problem as was extensively studied in~\cite{mainPaper}. Similarly, for Wasserstein regression models, the problem turns into the computation of a weighted Wasserstein barycenter problem. 

It follows from~\cite[Theorem 2.4]{barycenterEquation} that the Wasserstein barycenter of a set of zero-mean Gaussian random distributions $\{\mathcal{N}(0,\Sigma_i)\}_{i=1}^n$ each with non-negative weights $\lambda_{i}$ such that $\sum_{i = 1}^{n} \lambda_{i} = 1$ is a zero-mean Gaussian distribution with covariance matrix defined by the following implicit equation $S = \sum_{i = 1}^{n} \lambda_{i}(S^{\frac{1}{2}}\Sigma_{i}S^{\frac{1}{2}})^{\frac{1}{2}}$, which has been shown to be well-defined \cite{barycenterEquation}. 
\begin{theorem}[Theorem 4.2 in~\cite{barycenterEquation}]\label{thm_fixed}
    Let $\{L_{i}^{\dagger}\}_{i=1}^n$ be a set of $d\times d$ positive semidefinite matrices, with at least one of them positive definite. For a positive definite $S_0$, and a set of non-negative weights $\{\lambda_i\}_{i=1}^n$, with $\sum_{i=1}^n \lambda_i =1$, define
    \begin{align*}
S_{t+1} = S^{-\frac{1}{2}}_{t}\left( \sum_{i = 1}^{n} \lambda_i(S^{\frac{1}{2}}_{t}L_{i}^{\dagger}S^{\frac{1}{2}}_{t})^{\frac{1}{2}}\right)^{2}S^{-\frac{1}{2}}_{t}, \ t \geq 0.
\end{align*}
Then, $W_{2}(\mathcal{N}(0, S_{t}), \mathcal{N}(0, S)) \rightarrow 0 $ as $t \rightarrow \infty$.
    \end{theorem}

Theorem~\ref{thm_fixed} implies that as $t$ grows, $S_t$ approaches the covariance of the weighted barycenter of the set of Gaussian distributions. However, by representing each graph $G_i$ as a multivariate Gaussian $\nu^{G_{i}} = \mathcal{N}(0, L_{i}^{\dagger})$, graph Laplacians have a zero eigenvalue~\cite{andreotti2018multiplicity}; thus, the conditions in~\cite[Theorem 4.2]{barycenterEquation} do not hold because pseudo-inverses of these Laplacians also have this zero eigenvalue~\cite{pseudoInverse}, meaning that our Gaussians will have non-invertible covariance matrices and thus will be degenerate. The degeneracy issue of graph Laplacians can be solved by considering the modified fixed-point iteration proposed in~\cite{haasler2023bureswasserstein} that shifts the covariances before iteration
\sloppy
\begin{align*}
S_{t+1} {=} S^{-\frac{1}{2}}_{t}\left( \sum_{i = 1}^{n} \lambda_i\left(S^{\frac{1}{2}}\left(L_{i}+\frac{1}{k} \mathbf{1}_{k^2}\right)^{-1}S^{\frac{1}{2}}_{t}\right)^{\frac{1}{2}}\right)^{2}S^{-\frac{1}{2}}_{t}
\end{align*}
and then shifts the resulting barycenter back. Here, $|V| = k$. Their convergence result is built on the following proposition
\begin{proposition}[Propositon 3.4 in~\cite{haasler2023bureswasserstein}]
    The Bures-Wasser-stein barycenter of the set of graph Laplacians $\{L_{i}^{\dagger}\}_{i=1}^n$ is also a Bures-Wasserstein barycenter for the set of graph Laplacians $\{L_{i}^{\dagger} + (1/|V|) \mathbf{1}_{|V|^2}\}_{i=1}^n$.
\end{proposition}

Another approach to tackle the non-degeneracy is to consider Entropy Regularized Wasserstein distances and their barycenters~\cite{entropicBarycenter}, where they propose the  fixed-point iteration for an arbitrary small $\varepsilon>0$ as
\begin{align}\label{eq:ent_bar}
     S = \frac{\varepsilon}{4}\sum_{i = 1}^{n} \lambda_i \left(-I + \left(I + \frac{16}{\varepsilon^{2}}S^{\frac{1}{2}}L_{i}^{\dagger}S^{\frac{1}{2}} \right)^{\frac{1}{2}} \right).
\end{align}
Although the fixed-point iteration for $\varepsilon=0$ is known to converge, the authors in~\cite{entropicBarycenter} point out that it is still an open question whether~\eqref{eq:ent_bar} converges in other cases. The evidence in our numerical analysis suggests a positive answer. Additional approaches for the computation of Bures-Wasserstein barycenters of positive-semidefinite matrices, and in turn solving the barycenter of graphs, can be found in~\cite{zheng2023barycenter}.

The proposed network regression algorithm is described as follows\footnote{An alternative algorithm for the computation of the barycenter in the Bures-Wasserstein space can be found in \cite{haasler2023bureswasserstein}}

\begin{algorithm}
\caption{Entropy-regularized Wasserstein-Based Network Regression}\label{alg:cap}
\begin{algorithmic}[1]
\Require $x \in \mathbb{R}^d$, $\varepsilon \geq 0$, $\{(X_i,G_i)\}$, for $i=1,\cdots n$.
\State $\bar{X} = n^{-1} \sum_{i = 1}^{n}X_{i}$, $\hat{\Sigma} = n^{-1}\sum_{i = 1}^{n}(X_{i} - \bar{X})(X_{i} - \bar{X})^{T}$ For $i=1,\cdots,n$:
\State $L_{i}^{\dagger}$ is pseudo-inverse of Laplacian $L_{i}$ of graph $G_{i}$
\State $s_{iG}(X_i,x) =  1 + (X_{i} - \bar{X})^{T}\hat{\Sigma}^{-1}(x - \bar{X})$
\State \textbf{Solve for $S$}
\begin{align*}
     S = \frac{\varepsilon}{4}\sum_{i = 1}^{n}  s_{iG}(X_i,x) \left(-I + \left(I + \frac{16}{\varepsilon^{2}}S^{\frac{1}{2}}L_{i}^{\dagger}S^{\frac{1}{2}} \right)^{\frac{1}{2}} \right)
\end{align*}
\Return  Graph $G(x)$ with Laplacian $L(x) = S^{\dagger}$.
\end{algorithmic}
\end{algorithm}



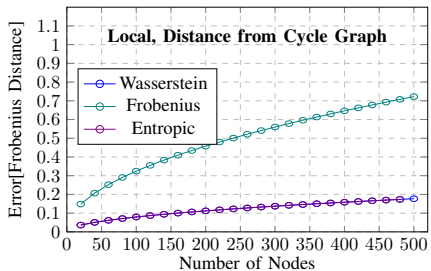
\begin{figure}[t!]
\centering
\begin{tikzpicture}[yscale = 0.7, xscale = 0.7]
\begin{axis}[
    title={\bf{Local, Distance from Cycle Graph}},
    xlabel={Number of Nodes},
    ylabel={Error[Frobenius Distance]},
    yscale = 0.75,
    xmin=0, xmax=520,
    ymin=0, ymax=1.2,
    xtick={0, 50, 100, 150, 200, 250, 300, 350, 400, 450, 500},
    ytick={0, 0.1, 0.2, 0.3, 0.4, 0.5, 0.6, 0.7, 0.8, 0.9, 1.0, 1.1},
    legend pos=north west,
    ymajorgrids=true,
    xmajorgrids=true,
    grid style=dashed,
]
\addplot[
    color=blue,
    mark=o,
    ]
    coordinates {
    (20,0.03628212276761326)(40,0.05065700429016684)(60,0.06177274265231076)(80,0.07117310858688912)(100,0.07946918293929293)(120,0.08697752978886517)(140,0.09388732789521)(160,0.10032233046757782)(180,0.10636874372473343)(200,0.11208947003485219)(220,0.1175320762339096)(240,0.12273356751500468)(260,0.12772340523982237)(280,0.13252549904534705)(300,0.13715956965655715)(320,0.14164210955609371)(340,0.14598707768369976)(360,0.15020641298443826)(380,0.15431042140046103)(400,0.15830807245369236)(420,0.1622072299718088)(440,0.1660148339948379)(460,0.1697370459193648)(480,0.1733793655789667)(500,0.1769467266144332)
    };

\addplot[
    color=teal,
    mark=o,
    ]
    coordinates {
    (20,0.14813669905572516)(40,0.20682795290731762)(60,0.25221255640429086)(80,0.2905934413070806)(100,0.324465618151089)(120,0.3551232308203677)(140,0.383333658523852)(160,0.40960723042547836)(180,0.4342942310055221)(200,0.4576514844951803)(220,0.47987320671001393)(240,0.5011104888498609)(260,0.5214836230365352)(280,0.5410902455159026)(300,0.5600100586297422)(320,0.5783118602345961)(340,0.5960519797912041)(360,0.6132791201647196)(380,0.6300354248716007)(400,0.6463582275449763)(420,0.6622773824055598)(440,0.677823462837093)(460,0.6930209751627813)(480,0.7078927402988323)(500,0.7224580711458324)
    };

\addplot[
    color=violet,
    mark=o,
    ]
    coordinates {
    (20,0.03630395612237378)(40,0.05068745581558306)(60,0.06180985871895423)(80,0.07121585754905116)(100,0.07951690630223876)(120,0.0870297555419678)(140,0.093943696762169)(160,0.1003825587629061)(180,0.10643259811636005)(200,0.11215675571334384)(220,0.11760262619590224)(240,0.12280723761199112)(260,0.12780006814050518)(280,0.13260504220681393)(300,0.1372418928276636)(320,0.14172712168714588)(340,0.14607469609258808)(360,0.1502965624802891)(380,0.15440303281229759)(400,0.15840308140063578)(420,0.16230457769946402)(440,0.16611446824552634)(460,0.16983890991903608)(480,0.17348341650434781)
    };
    \legend{Wasserstein, Frobenius, Entropic}  

\node[] at (420, 74.153454967390051) {(500, 0.7224)};
\node[] at (420, 22.041478757080223) {(500, 0.1769)};
\end{axis}
\end{tikzpicture}
\caption{The Frobenius distance between the predicted graph for $x=5$ and ground truth trained on cycle graphs with an increasing number of nodes for Wasserstein, Frobenius, and Entropic Wasserstein regressors -- an extension of experiment in Figure~\ref{fig:1}. This demonstrates that the Wasserstein-based regressors outperform the Frobenius across networks of varying sizes. Moreover, the error growth is slower for the Wasserstein regressors,  suggesting their superior performance over large-scale networks. Note that the Wasserstein and Entropic Wasserstein outputs are indistinguishable.}
\label{fig:2}
\vspace{-0.5cm}
\end{figure}

\subsection{An open problem in affine combinations for positively curved spaces}

Note that in our case, Problem~\eqref{eq:emp_reg} can be understood as a barycenter problem where the weights are defined as $\lambda_i = s_{iG}(X_i,x)$ and determined by the regressors from the available data pairs and its computation is not trivial~\cite{zhou2023wasserstein,petersen2019frechet,fan2021conditional}. However, even though $\mathbb{E}[s_{iG}(X,x)]=1$, the weights as defined for $\hat{m}_{G}(x)$ can be negative. 

The generic structure of Problem~\eqref{eq:emp_reg} with possibly negative weights that add up to one can be understood as the computation of the Riemannian center of mass~\cite{huning2019convergence}, which arises from the subdivision schemes. The existence of a unique minimizer for Fr\'echet means in Riemannian manifolds for Cartan-Hadamard (nonpositive sectional curvature) is well understood~\cite[Theorem 6]{huning2019convergence}. However, the Wasserstein space is a non-negatively curved metric space~\cite[Section 7.3]{ambrosio2005gradient}, and only the existence of local minimizers can be guaranteed~\cite{sander2016geodesic}.

Some initial theoretical results on the existence and computation of Fr\'echet means on positively curved spaces have been recently proposed~\cite{huning2022convergence}. However, general theoretical results are still an open problem and lie outside of the scope of this work. Generally, the convergence of the previously described fixed-point iterations is not guaranteed~\cite{generalfrechetmeans}. However, in practice, convergence occurs for the studied scenarios. As discussed in~\cite{fan2021conditional}, the computation of Wasserstein barycenters with possibly negative weights, which turns the problem into an affine combination instead of a convex combination, remains an open problem. We propose these fixed-point iteration methods for computational purposes and leave a connection between Fr\'echet Means and General Fr\'echet Means as a future extension~\cite{wallner2020geometric}.



\section{Numerical Analysis}\label{sec:numerics}

\begin{figure}[t!]
    \centering    \includegraphics[width=0.45\textwidth]{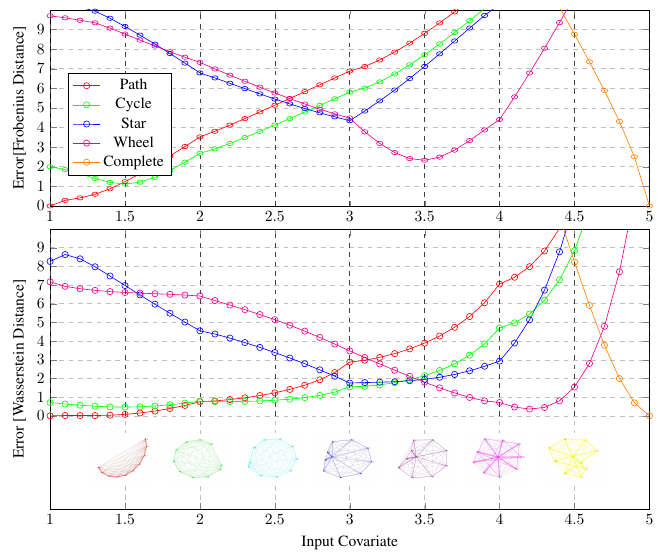}
    \caption{We train regressors over 5 input and response pairs $\{X_i, G_i\}_{i = 1}^5$ respectively. The covariate $X_i$ is an integer from 1 to 5, and $G_i$ is a named graph in the legend ordered by increasing connectivity, e.g., $X_1 = 1$ and $G_1$ is the path graph. We output the graphs predicted with the Wasserstein regressor for each 0.5 step between 1 and 5 on the x-axis. Each line plots the error between a named graph and interpolations over 0.1 steps for the Frobenius (top) and Wasserstein (bottom) regressors.
    }
    \label{fig:3}
\end{figure}

This section shows metric comparisons over synthetic and real-world graphs with various topologies. 

Initially, following the example presented in Figure~\ref{fig:1}, Figure~\ref{fig:2} shows the prediction error with respect to the Wasserstein, Frobenius, and Entropy-regularized Wasserstein distances for the case where $x=5$, as the number of nodes in the graph grows. Wasserstein-based regressors have a smaller prediction error and better error scalability. Appendix~\ref{app:B} shows additional results that provide evidence of the improved performance of $d_{W}$ over $d_{F}$ for various graph topologies, sizes, and regression tasks.

\textbf{Naive Interpolation of Topologies:} We consider $5$ feature and graph pairs: path, cycle, star, wheel, and complete graphs, each with $10$ nodes and corresponding integer covariate from $1$ to $5$ in order of increasing connectivity, e.g., $X_1 = 1$ and $G_1$ is the path graph.  

Figure~\ref{fig:3} shows the distance between the Frobenius and Wasserstein-based regressors' predictions, computed following~\eqref{eq:ent_bar}, and the five topologies used. Interpolating sample graphs with $d_{W}$ is more accurate as output graphs maintain smaller distances to graphs in the sample space. For example, the pink plot shows the distance between the predicted and wheel graphs. It is expected that a minimal value occurs at $x=4$. The prediction error for Wasserstein-based regressors is smaller than the Frobenius one.    
More instances of interpolated graph outputs with the Wasserstein distance are shown in Figure~\ref{fig:graph-interpolations}, and results for each named graph can be found in Appendix~\ref{app:c}.

\begin{figure}
    \centering
    \includegraphics[width=0.225\textwidth]{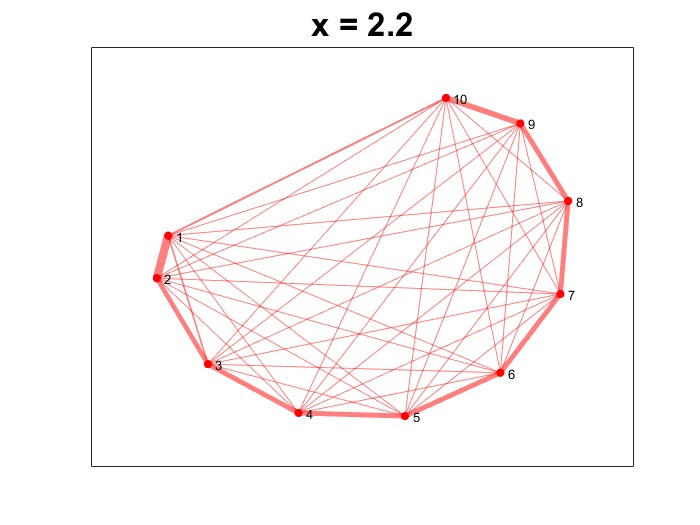} 
    \includegraphics[width=0.225\textwidth]{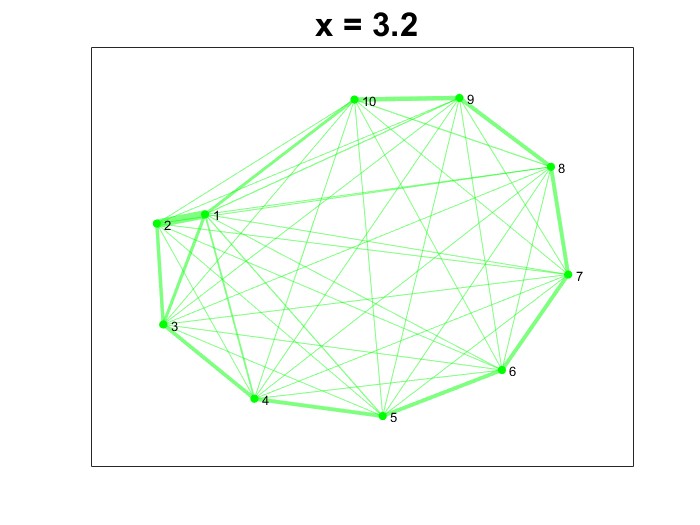}
    \includegraphics[width=0.225\textwidth]{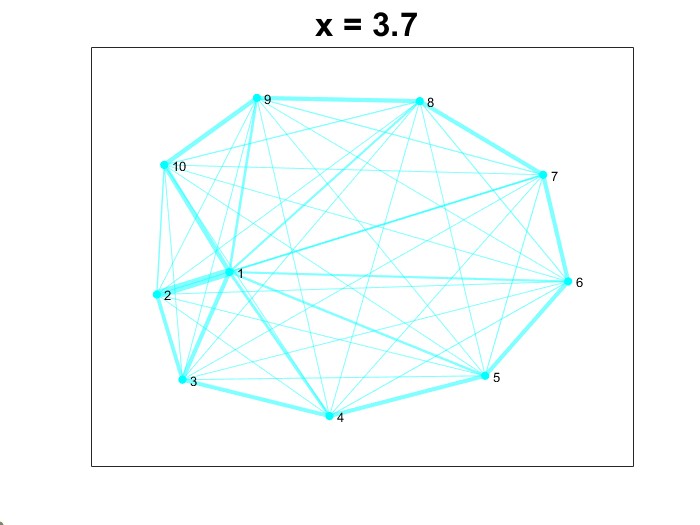}
    \includegraphics[width=0.225\textwidth]{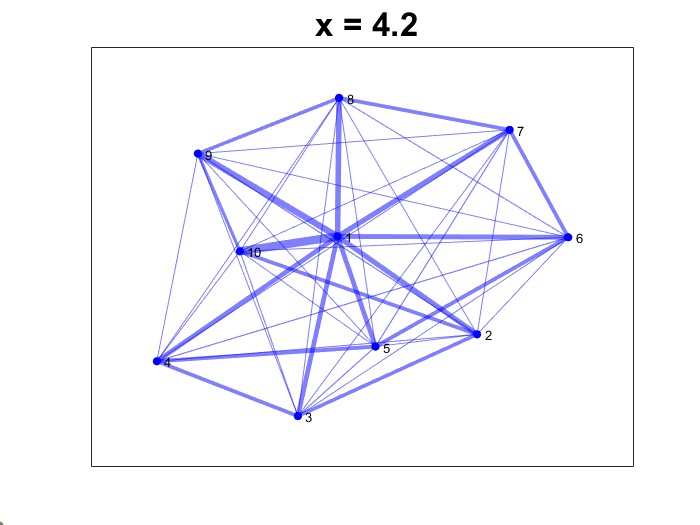}
    \caption{Additional interpolated graphs from the Wasserstein regressor in Fig~\ref{fig:3}, with inputs of 2.2, 3.2, 3.7, and 4.2.}
    \label{fig:graph-interpolations}
    \vspace{-.3cm}
\end{figure}

\begin{figure}
     \centering
     \begin{subfigure}[b]{0.235\textwidth}
         \centering
         \includegraphics[width=\textwidth]{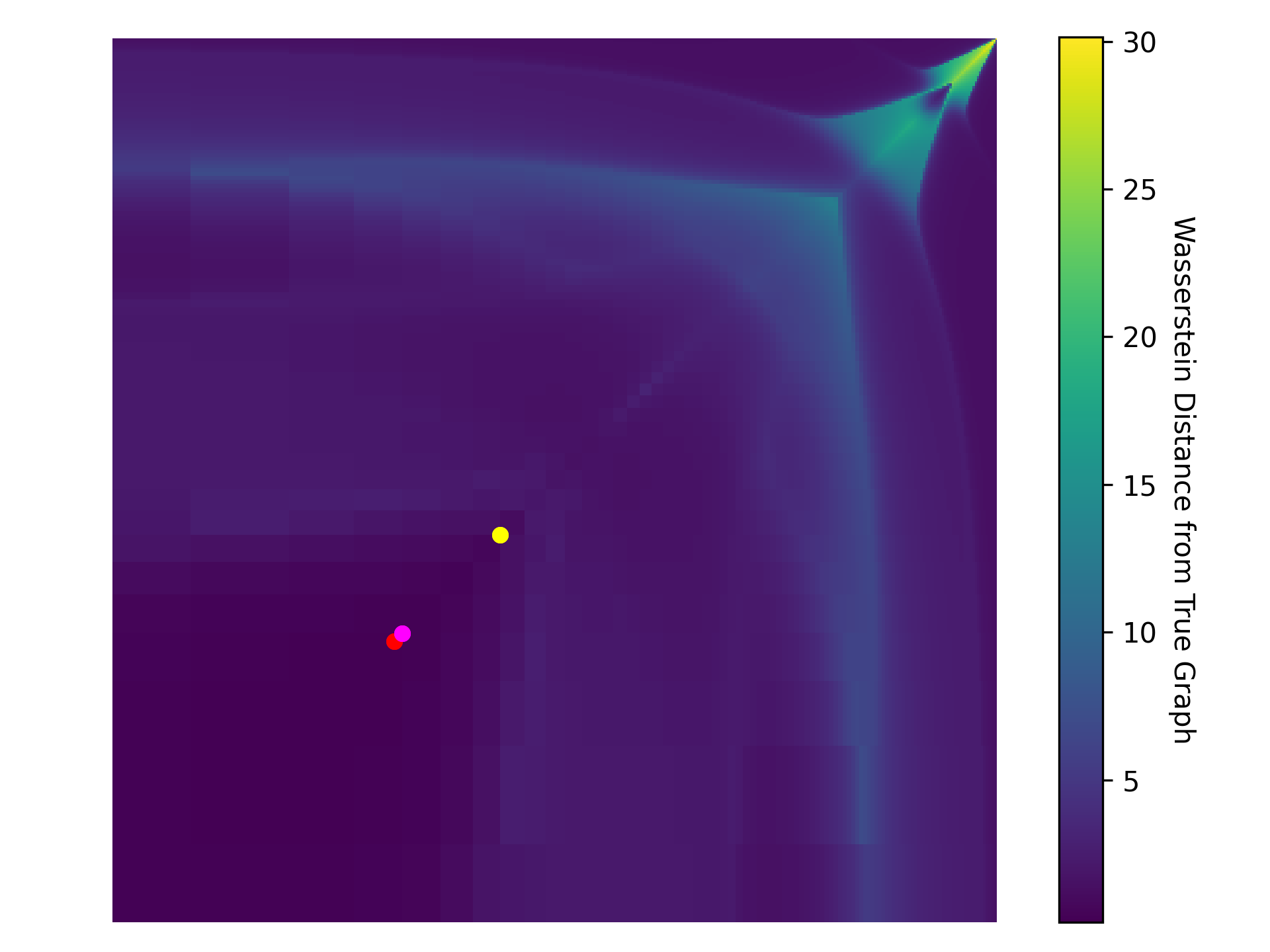}
         \caption{Distance from Wasserstein regressors to cycle}
         \label{2-cycle}
     \end{subfigure}
     \begin{subfigure}[b]{0.235\textwidth}
         \centering
         \includegraphics[width=\textwidth]{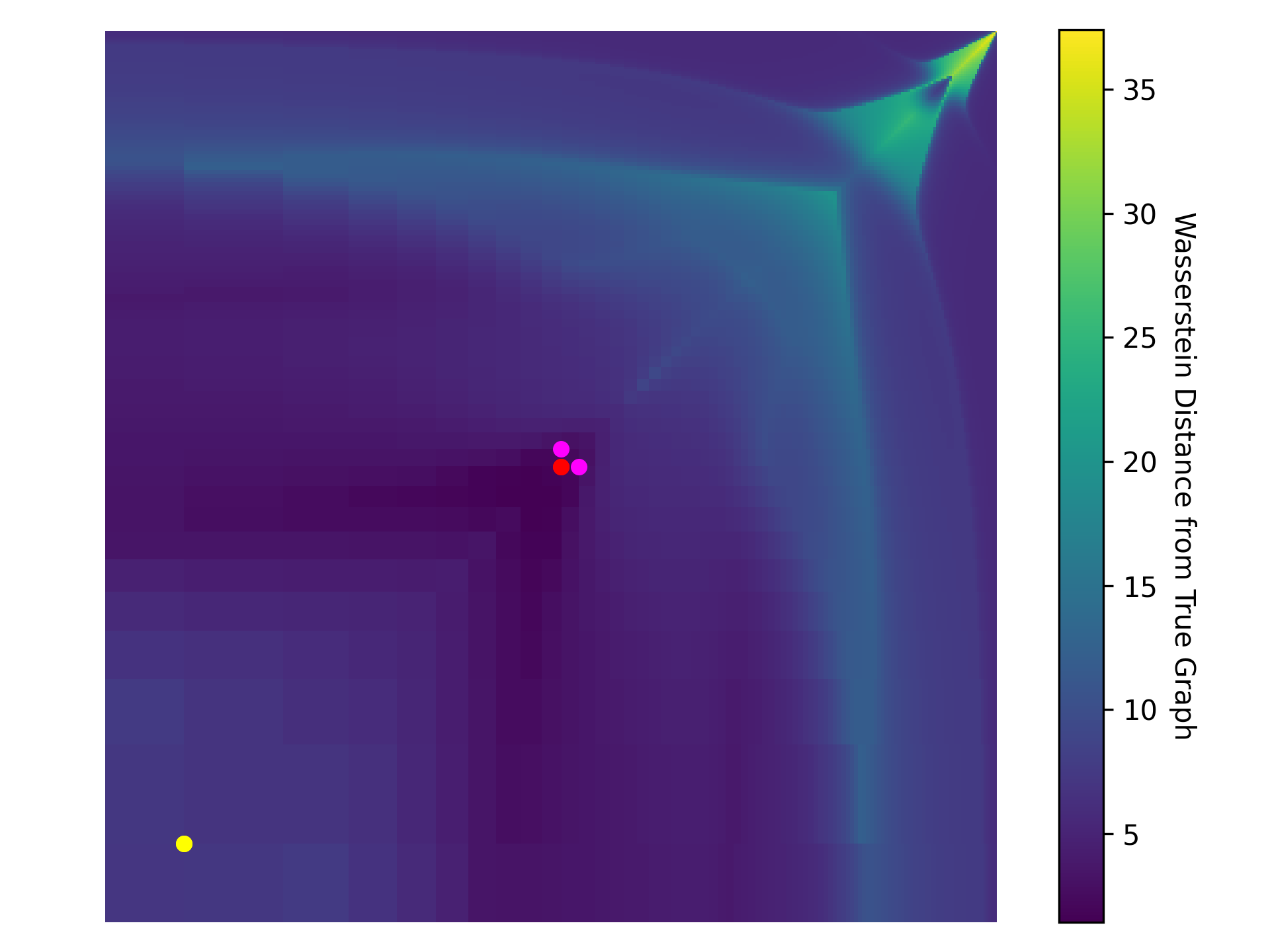}
         \caption{Distance from Wasserstein regressors to star}
         \label{6-regular}
     \end{subfigure}
     \begin{subfigure}[b]{0.235\textwidth}
         \centering
         \includegraphics[width=\textwidth]{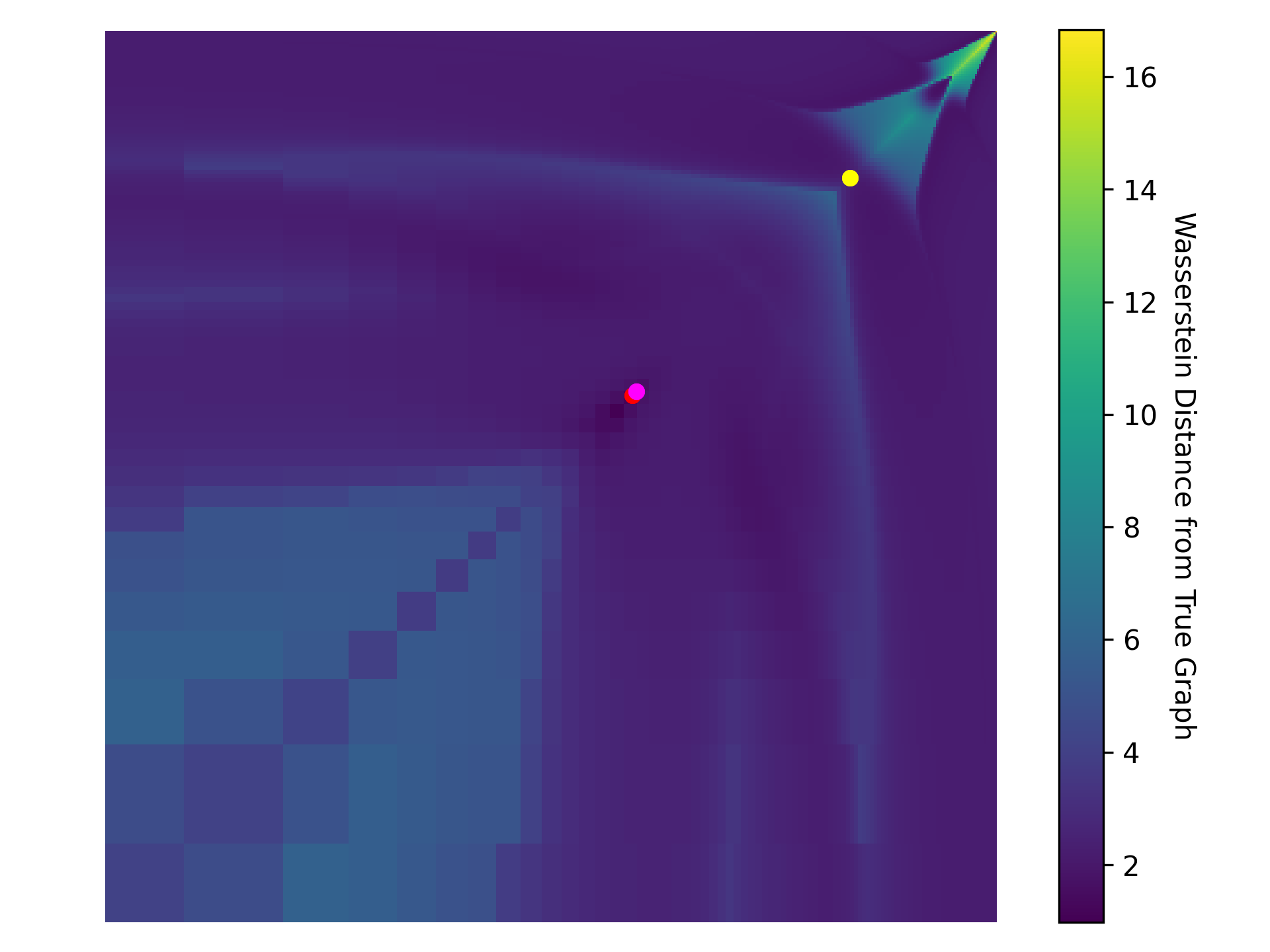}
         \caption{Distance from Wasserstein regressors to wheel}
         \label{wheel}
     \end{subfigure}
     \begin{subfigure}[b]{0.235\textwidth}
         \centering
         \includegraphics[width=\textwidth]{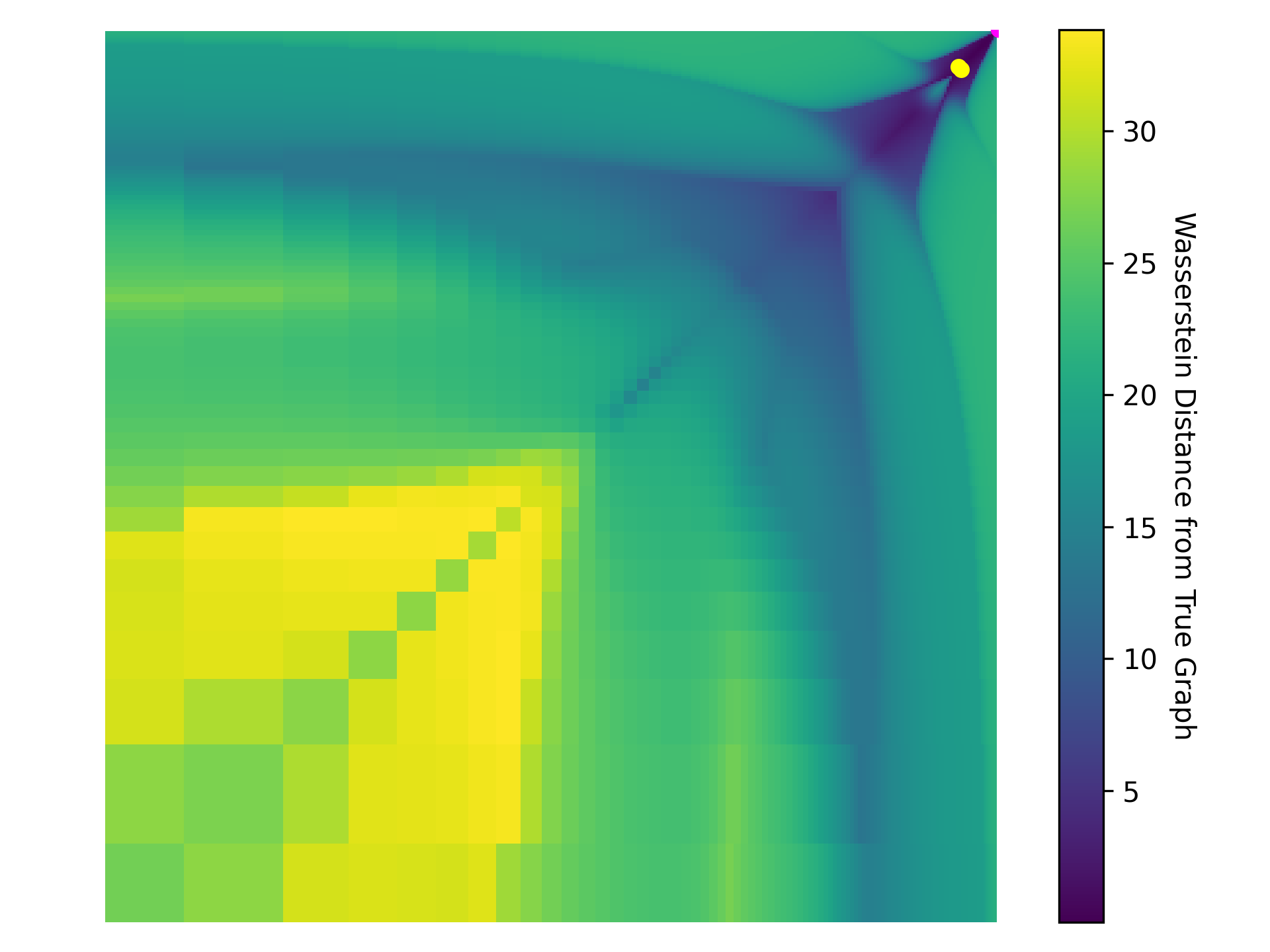}
         \caption{Distance from Wasserstein regressors to complete}
         \label{2-star}
     \end{subfigure}
\caption{Heatmaps representing the distance from the Wasserstein regressor to true graphs for cycle (a), star (b), wheel (c), and complete (d), with minimums for Frobenius (yellow), Wasserstein (magenta), and ground truth (red) as points. For all of these heatmaps, the $x$-axis is the value of the second smallest eigenvalue, and the $y$-axis is the value of the 3rd smallest eigenvalue, both being logarithmic and ranging from $1$ to $100$}
\label{fig5}
\vspace{-0.5cm}
\end{figure}

\textbf{Spectrum-based Interpolation of Topologies:} We extend our previous experiment by computing regressors over the spectral properties of the graph Laplacians instead of the (arbitrary) integers assigned before. We consider a set of $n=12$ named graphs: path, star, cycle, wheel, complete, dumbbell, lollipop, Pentagonal Prism, two-star (two stars connected by their center), and $4$, $6$, and $8$ regular. Each graph has $10$ nodes, i.e., $|V|=10$, where the covariate $X_i \in \mathbb{R}^2$ is a vector containing the corresponding second and third smallest eigenvalues of the graph Laplacians. We computed the regressor in~\eqref{eq:emp_reg} to generate predicted graphs for $x \in (0,10]^2$. Figure~\ref{fig5} shows the distance between the predicted graphs and four named graphs used in the dataset. 
The eigenvalues of our graphs correspond to the log base axes of the heatmaps, with the closest predicted graph to the true graph in magenta and the true graph in red.

Figure~\ref{fig5} shows that using the spectral properties of the sample graphs leads to precise predictions, having output graphs occurring near the true graphs and avoiding graphs of different connectivities. 


\textbf{Large-Scale Real Data:} We validate the performance of the Wasserstein-based methods by comparing its performance on the network inference task of taxi usage in response to the number of COVID-19 cases in Manhattan, studied for Frobenius-based regressors in~\cite{mainPaper}. We reproduce the original experiment and show that the Wasserstein regressor outperforms the Frobenius regressor.

We take rider data from~\cite{taxiData}, including the number of passengers, the pickup and dropoff location, and the day the trip occurs. For a given day, we construct a graph where each node represents one of $13$ Manhattan regions, and the edge weights represent the number of riders traveling between regions. From these graphs, we compute $172$ graph Laplacians from April $12, 2020$, to September $30, 2020$. We then regress over these Laplacians in response to a binary weekend indicator, equaling $1$ if the day is a weekend and $0$ if not, and the daily number of COVID-19 cases in Manhattan~\cite{covidData}. We contrast the local Wasserstein regressor against the Frobenius regressor for a select date in Figure~\ref{fig:5}. One can see the Wasserstein regressor approximates edge weights closer to the ground truth than the Frobenius regressor, particularly in upper and lower regions of the network.

To quantitatively evaluate our model using all available data, we calculate the Fr\'echet version of the $R^{2}$ coefficient, which has similar interpretations of model fitness~\cite{petersen2019frechet}, and is defined for global models as $R^{2}_{\oplus} = 1 - {E[d^{2}(G, m_{G}(X))]}/{V_{\oplus}}$. Global results are $\hat{R}^{2}_{\oplus} = 0.433$ for the Power metric with $\alpha = 1$, $\hat{R}^{2}_{\oplus} = 0.453$ for the Power metric $\alpha = 1 /2$, and $\hat{R}^{2}_{\oplus} = 0.607$ for the Wasserstein metric. When using the method from~\cite{haasler2023bureswasserstein}, $\hat{R}^{2}_{\oplus} = 0.592$. 

\begin{figure}
\begin{center}
\begin{tikzpicture}[yscale = 0.75, xscale = 0.75]
\pgfplotsset{
width= 330,
height= 330
}
\begin{axis}[
    title={},
    xlabel={Covid Cases},
    xmode={log},
    ylabel={Weekend (1 if True)},
    xmin=35, xmax=280,
    ymin=-0.5, ymax=1.5,
    xtick={},
    ytick={0, 1},
    yscale = 0.7,
    legend pos=north west,
    ymajorgrids=true,
    xmajorgrids=true,
    grid style=dashed,
]
\node[] at (231, 0) {\includegraphics[width=1.5cm]{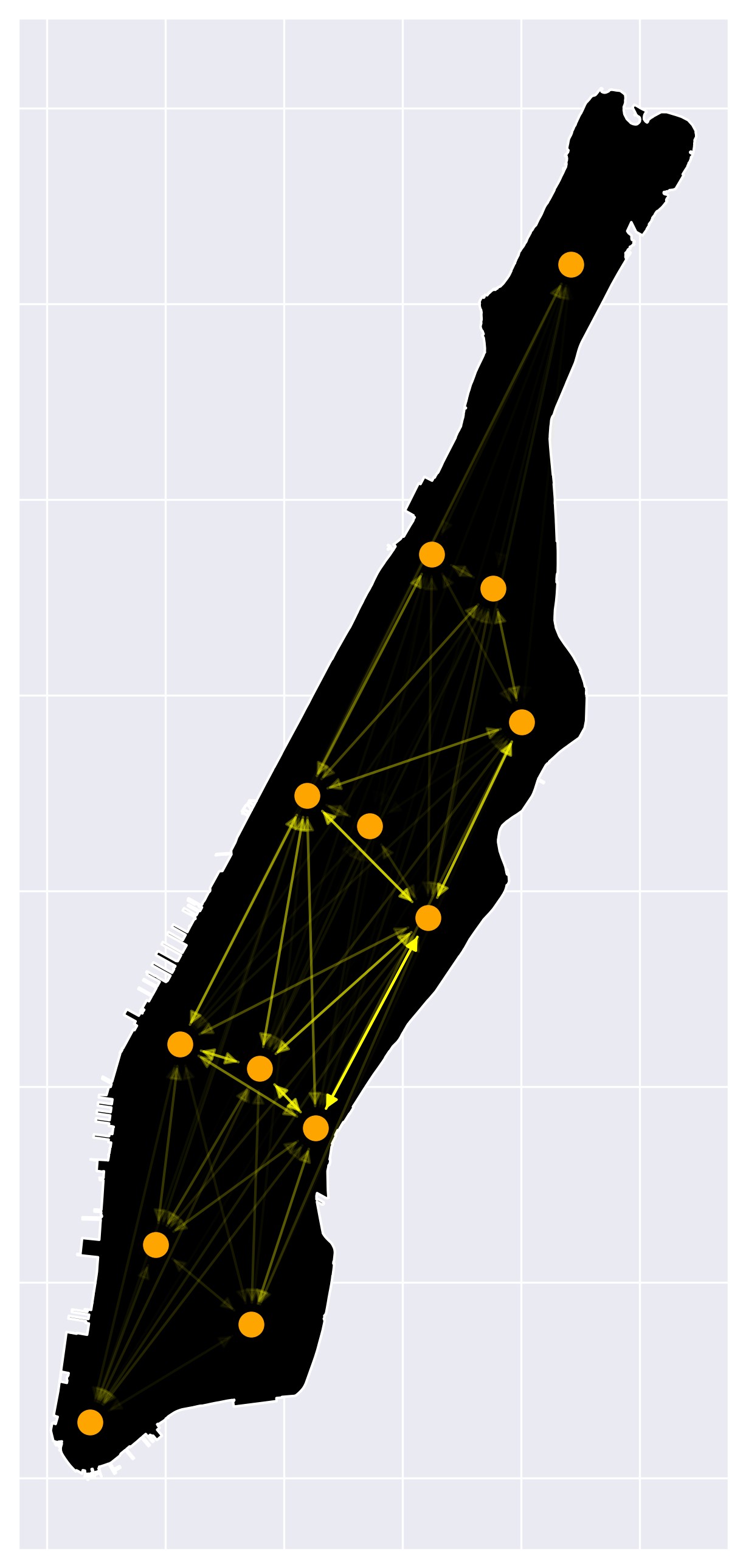}};
\node[] at (122, 1) {\includegraphics[width=1.5cm]{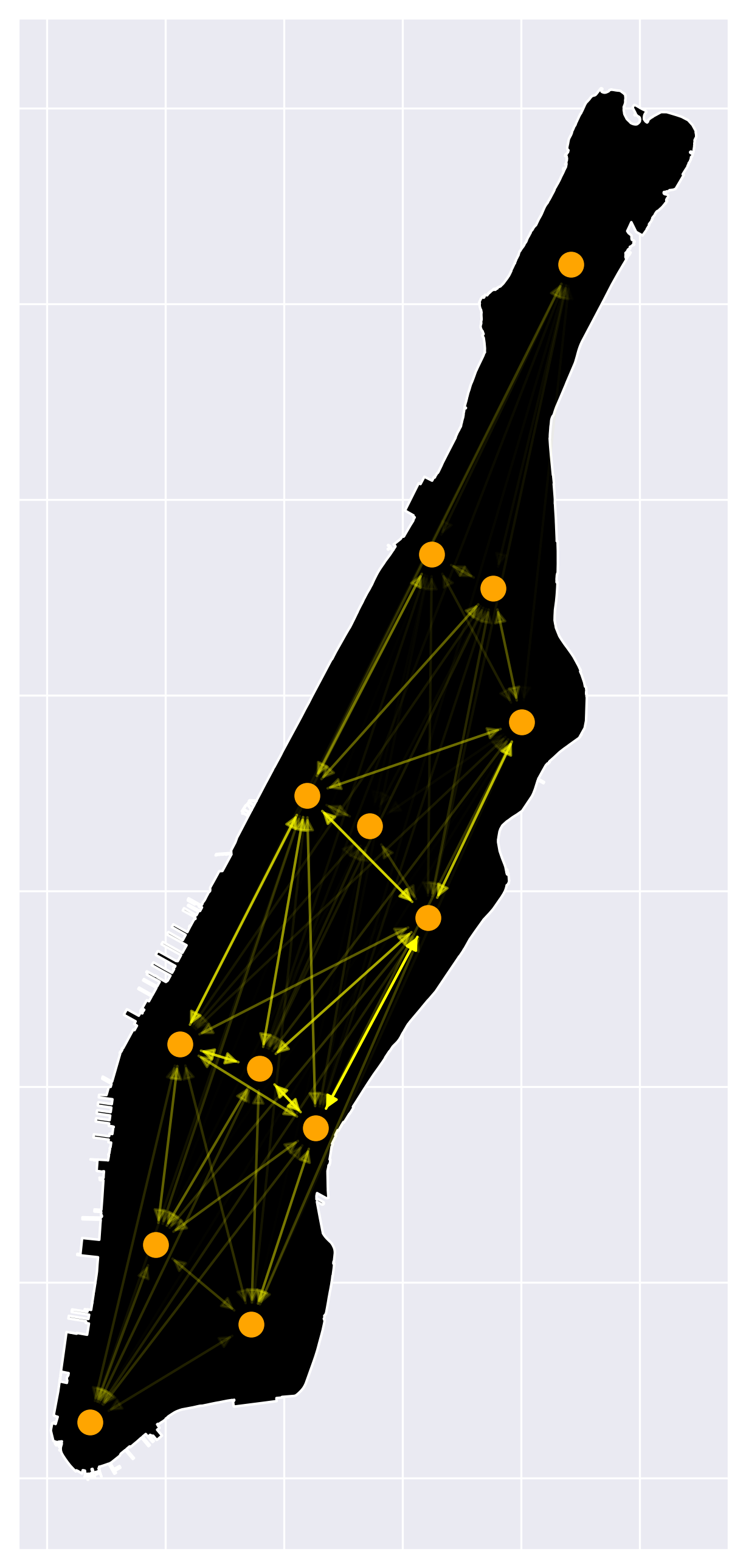}};
\node[] at (129, 0){\includegraphics[width=1.5cm]{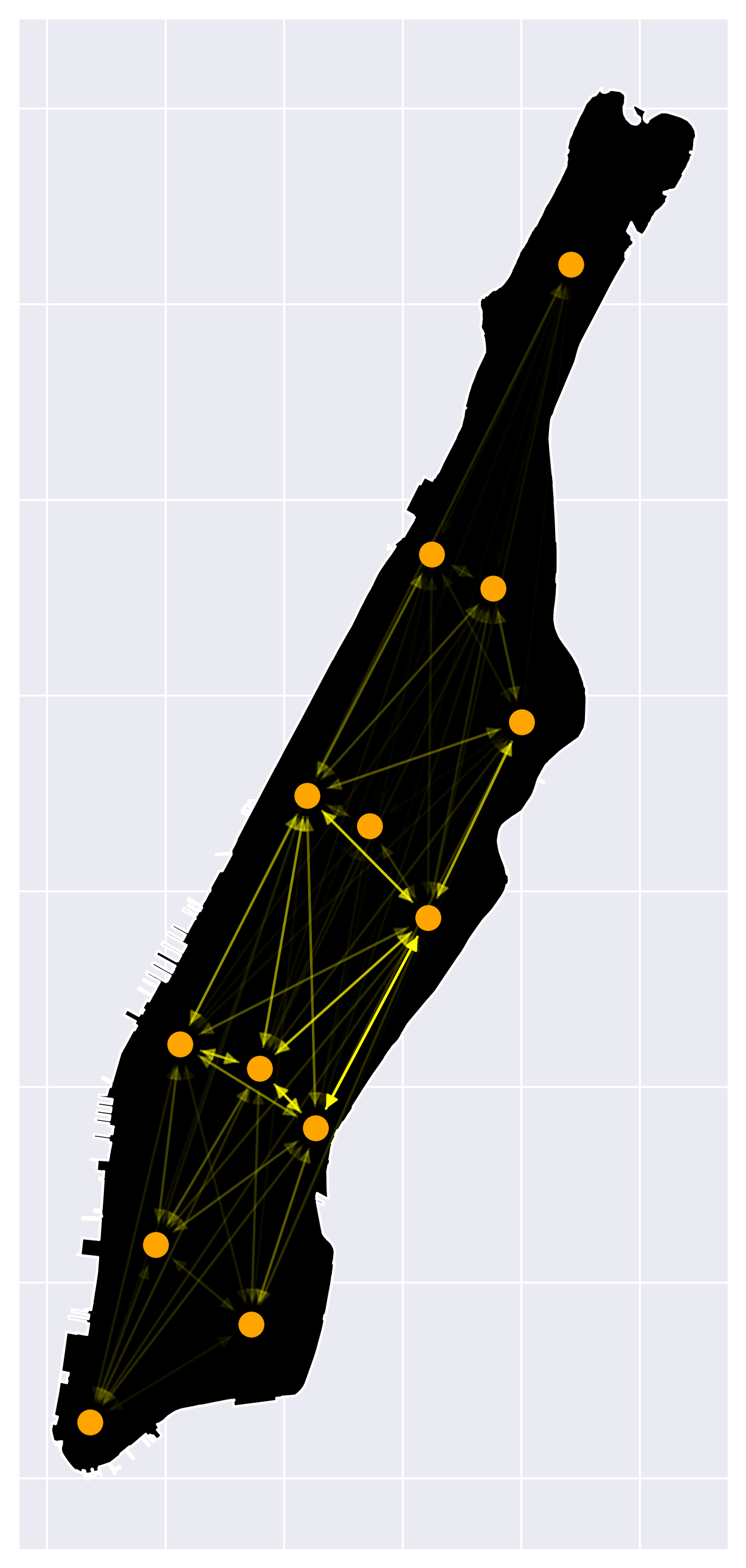}};
\node[] at (71, 1) {\includegraphics[width=1.5cm]{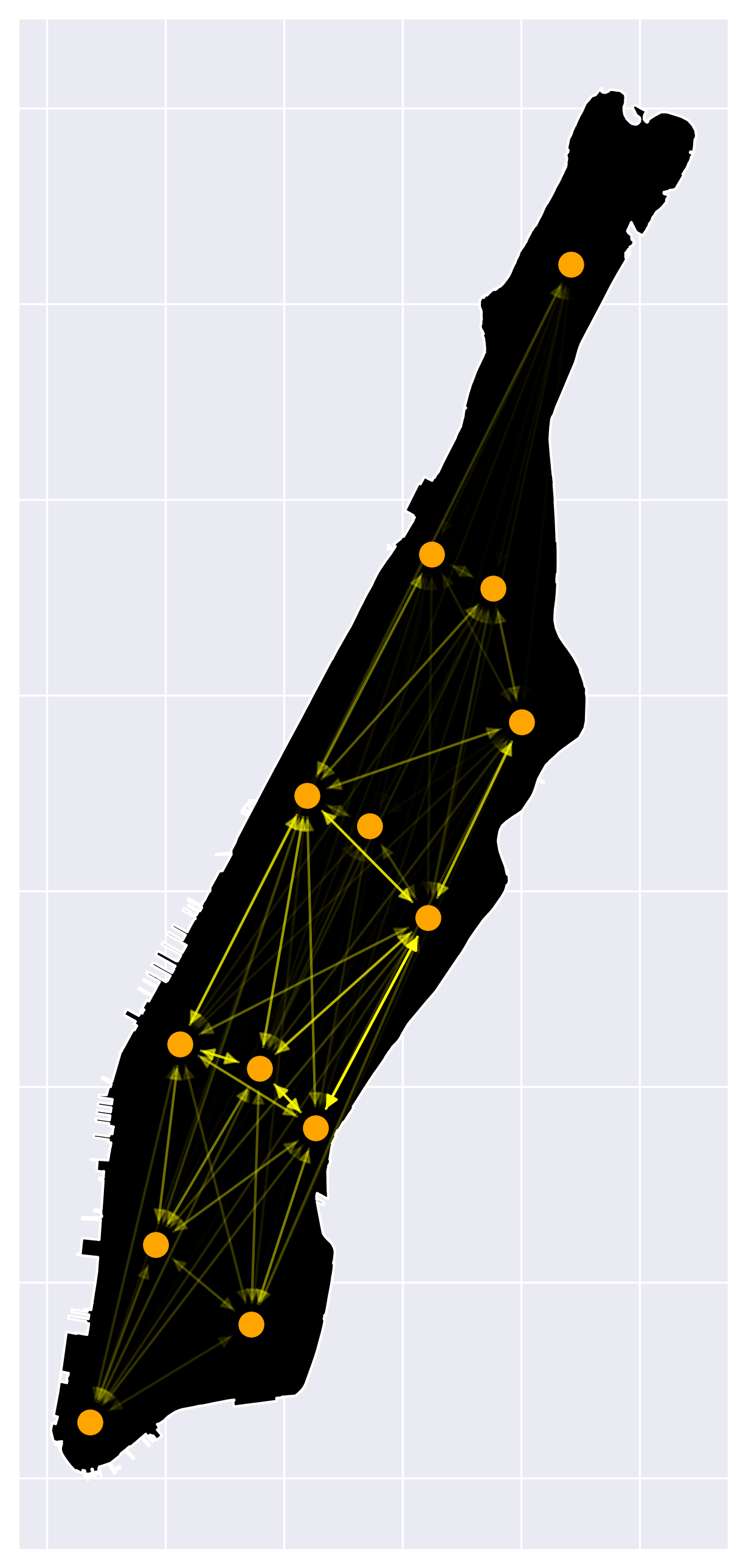}};
\node[] at (43, 1) {\includegraphics[width=1.5cm]{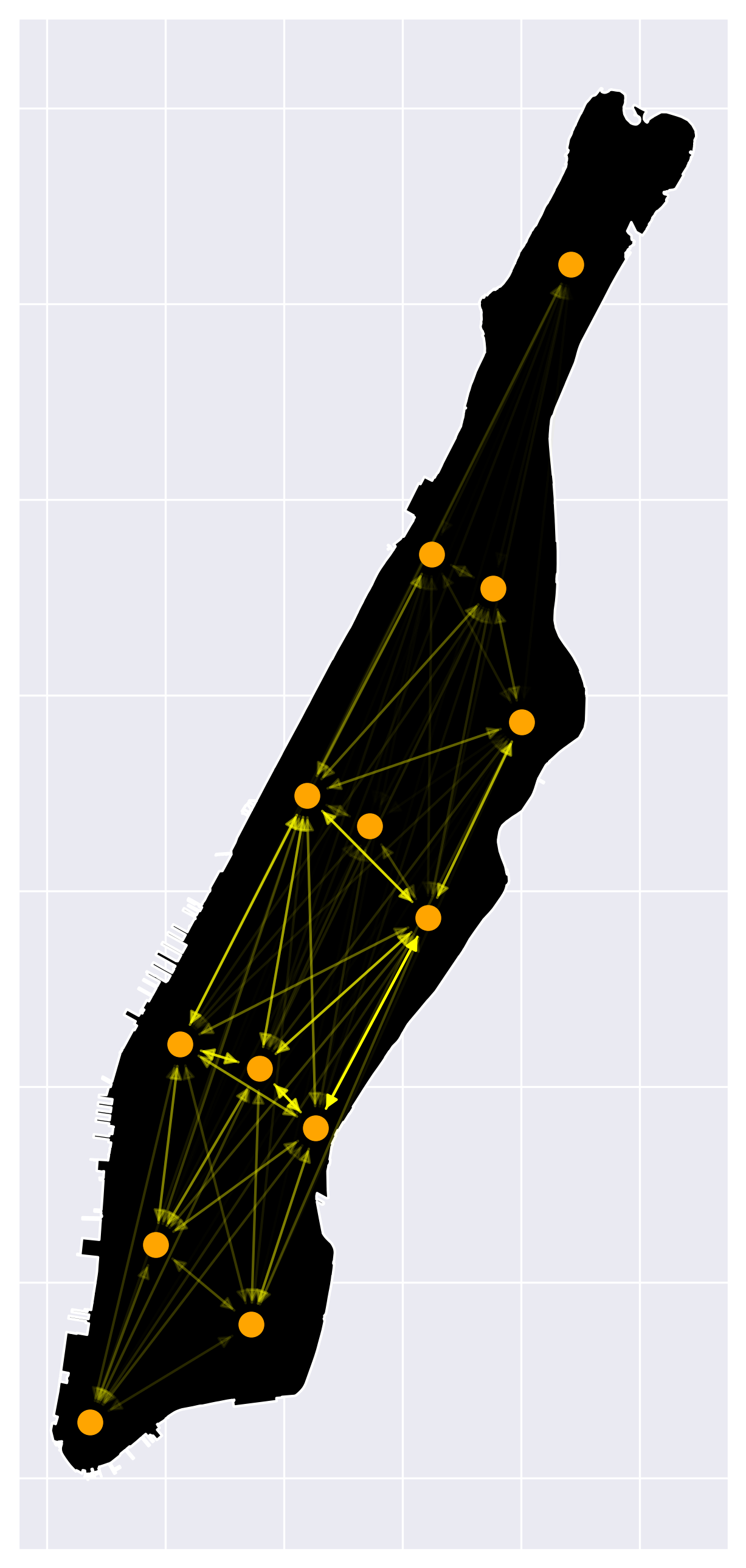}};
\node[] at (49, 0) {\includegraphics[width=1.5cm]{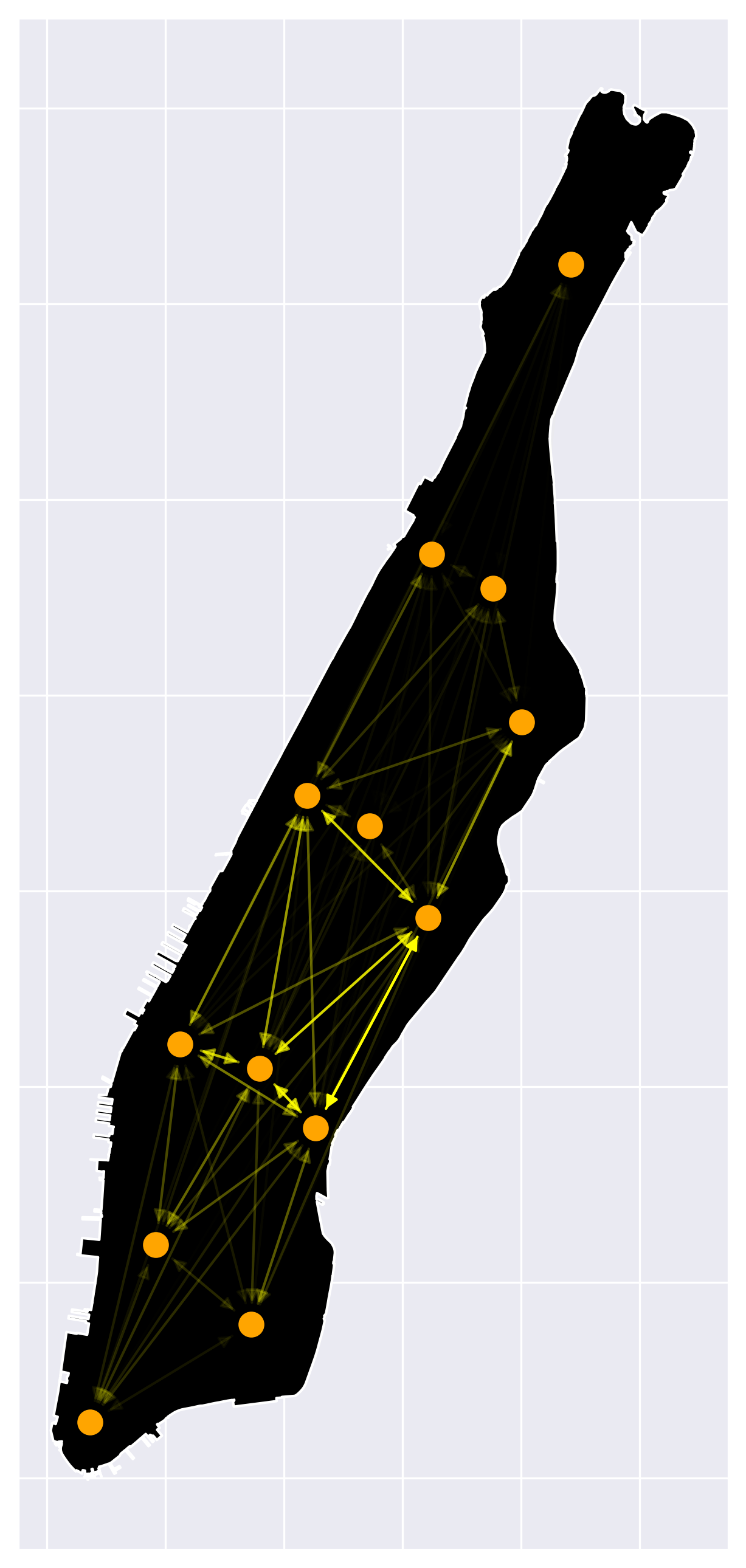}};
\end{axis}
\end{tikzpicture}
\end{center}
\caption{A plot of 6 different predictions using the Wasserstein regressor, where each output is a plot corresponding to its feature value, the x-axis representing the number of COVID cases and the y-axis representing a binary variable which equals $1$ if the day is a weekend and $0$ otherwise. The edge weight scales edge brightness, i.e., traffic volume between two regions. As the number of COVID cases drops, we observe the taxi traffic between regions increasing. Moreover, the traffic growth is typically greater on weekends for similar COVID cases.}
\vspace{-0.5cm}
\label{fig6}
\end{figure}

\begin{figure}
    \centering    \includegraphics[width=0.25\textwidth]{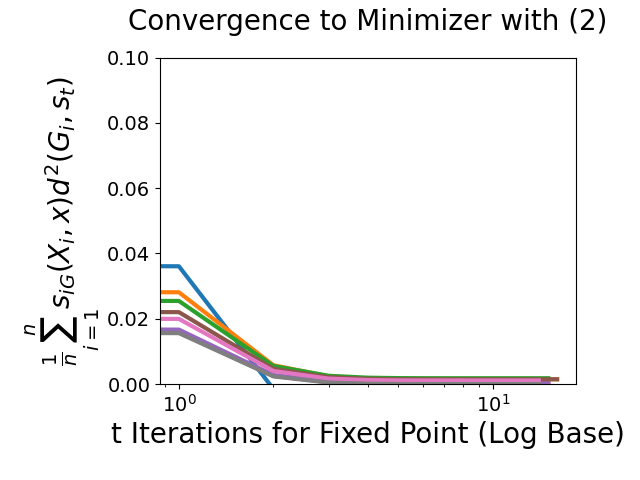}\includegraphics[width=0.25\textwidth]{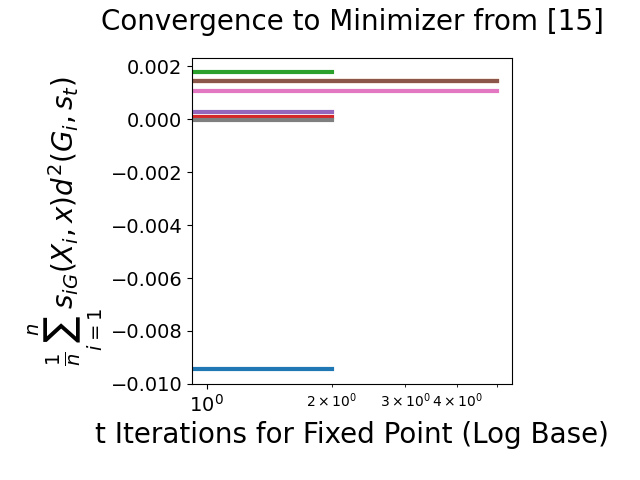}
    \caption{The number of iterations until convergence for our method on the left and existing methods ~\cite{haasler2023bureswasserstein} on the right. These iterations are graphed against the weighted sum of the distances between our output and the sample graphs, which is minimized through our regression.}
    \label{fig7}
    \vspace{-0.2cm}
\end{figure}

When computing this $R^{2}_{\oplus}$, we want to be sure that our methods are computationally efficient. In Figure~\ref{fig6}, we see the layout of 6 different inputs in their feature space. In Figure~\ref{fig7}, we see the number of iterations it takes for the iterates generated in \eqref{eq:emp_reg} to converge to its minimizer for these inputs. In all these cases, for both our methods and the methods in~\cite{haasler2023bureswasserstein}, we see convergence in less than 20 iterations.

Additionally, we use 10-fold cross-validation~\cite{mainPaper} to compute the mean square prediction error (MSPE) with both the Frobenius and Wasserstein metrics. Prediction can occur with either distance, but error computation should be consistent to have comparable accuracy of results. Thus, we have two results: error of Frobenius and Wasserstein predictions measured with the Frobenius distance and error measured with the Wasserstein distance. When averaging over 100 iterations, the MSPE can be seen in Table~\ref{table1}, leading to two main conclusions. 
First, even when measuring error with the Frobenius distance, the Wasserstein metric is still an improvement over the power metric, which is an adaptation of the Frobenius distance~\cite{mainPaper} that we would assume to have a smaller error for a similar metric. Secondly, when we compute the error with Wasserstein distances, we see a large decrease, showing the extent of our improved predictions when measuring error with the distance we model with.

\begin{table}
    \centering
    \caption{Accuracy Relative to Frobenius, smaller is better.}
    \begin{tabular}{|c|c|}
        \hline
        Distance Used & \% MSPE of Frobenius \\
        \hline
        Power Metric $d_{F,\alpha}$ $\alpha=0.5$ & 96.4\% \\
        Wasserstein (Prediction) Frobenius (Error) & 95.995\% \\
        Wasserstein (Prediction) Wasserstein (Error) & 86.375\% \\
        \hline
    \end{tabular}
    \label{tab:accuracy}
    \vspace{-0.3cm}
\label{table1}
\end{table}

\begin{figure}
    \vspace{-0.5cm}
    \centering \includegraphics[width=0.38\textwidth]{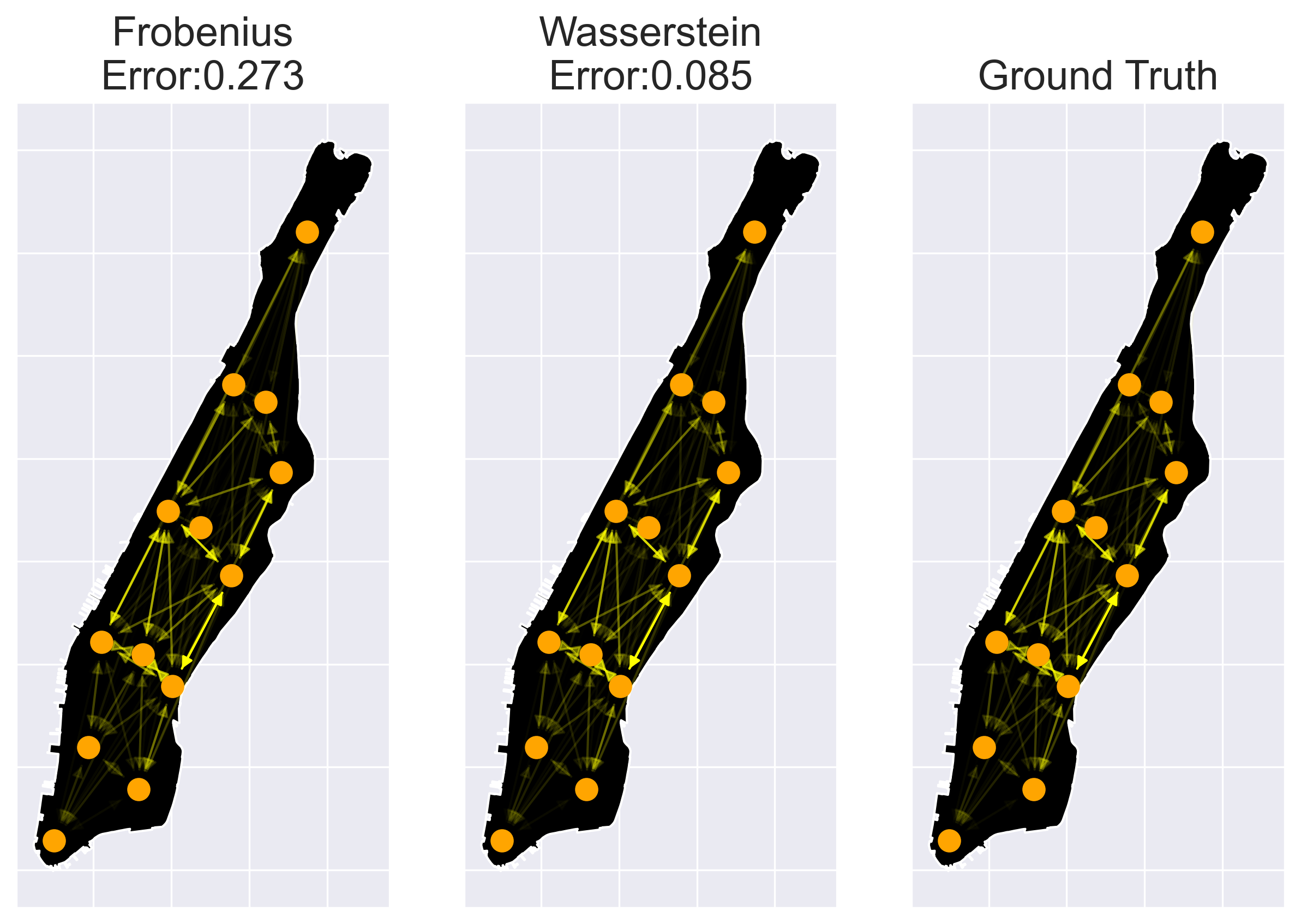}
    \caption{Graph prediction for Taxi Cab ridership on April 12, 2020, with local Frobenius (left), Wasserstein (center), and true network (right). Wasserstein regressor predicts edge weights more accurately, as seen by their Wasserstein distance from the true network. Edge coloring is scaled by edge weight.}
    \label{fig:5}
\end{figure} 

\section{Conclusion}\label{sec: conclusion}
We provided evidence for the superior performance of Wasserstein distances over the Frobenius norm in graph regression problems via experiments focused on network size, network structure, network variability, and analysis of real-world networks. In all of these instances, the global and local variants of the Wasserstein models obtain greater accuracy relative to their Frobenius counterparts. While our models generalize to arbitrary metric spaces, the computation accuracy unique to Wasserstein Regression compared to other graph prediction methods is vital to its applicability. We hope to motivate future efforts in network prediction by applying the Wasserstein metric to a wider breadth of real-world systems that include data sets with graphs of differing $|V|$. This extension would utilize the Gromov-Wasserstein distance, which has been used in graph prediction previously~\cite{brogatmotte2022learning} and generalizes the Wasserstein distance over graphs of differing sizes. By further investigating the Riemannian center of masses in the space of Gaussians measured with the Wasserstein distance, further research can ensure the convergence of these General Frech\'et Means. Such extensions would theoretically support our methods over networks with a larger breadth of variance, cementing our methods as essential in future graph prediction advancements.


\medskip





\vspace{-0.5cm}
\appendix

\subsection{Metric Regression Local Model Definition}\label{app:A}

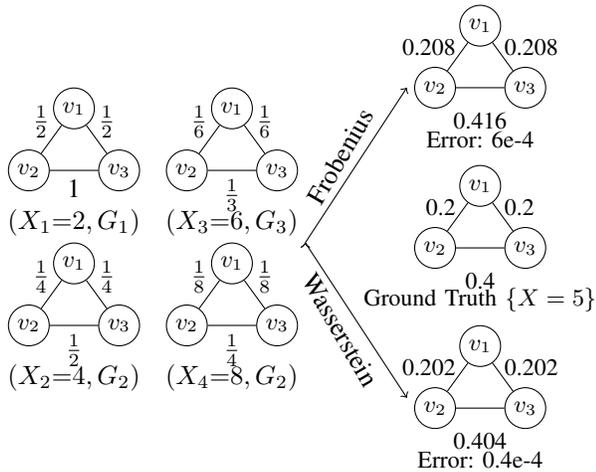
\begin{figure}[h!]
\centering
\begin{tikzpicture}[xscale = 0.3, yscale = 0.55]
\begin{scope}[every node/.style={circle, inner sep=2pt}, minimum size = 0.1em]
\node[draw] (a1) at (0, 2.5) {\small $v_{1}$};
\node[draw] (a2) at (-2, 1.0) {\small $v_{2}$};
\node[draw] (a3) at (2, 1.0) {\small $v_{3}$};
\node (a4) at (0, -0.25) {$(X_1{=}2,G_{1})$};

\node[draw] (b1) at (0, -1.25) {\small $v_{1}$};
\node[draw] (b2) at (-2, -2.75) {\small $v_{2}$};
\node[draw] (b3) at (2, -2.75) {\small $v_{3}$};
\node (b4) at (0, -4) {$(X_2{=}4,G_{2})$};

\node[draw] (c1) at (7, 2.5) {\small $v_{1}$};
\node[draw] (c2) at (5, 1.0) {\small $v_{2}$};
\node[draw] (c3) at (9, 1.0) {\small $v_{3}$};
\node (c4) at (7, -0.25) {$(X_3{=}6,G_{3})$};

\node[draw] (d1) at (7, -1.25) {\small $v_{1}$};
\node[draw] (d2) at (5, -2.75) {\small $v_{2}$};
\node[draw] (d3) at (9, -2.75) {\small $v_{3}$};
\node (d4) at (7, -4) {$(X_4{=}8,G_{2})$};
\end{scope}
\begin{scope}[every edge/.style={draw=black}]

\draw  (a1) edge node[above left, yshift = -1mm, xshift = 1mm] {$\frac{1}{2}$} (a2);
\draw  (a1) edge node[above right, yshift = -1mm, xshift = -1mm] {$\frac{1}{2}$} (a3);
\draw  (a2) edge node[below] {1} (a3);

\draw  (b1) edge node[above left, yshift = -1mm, xshift = 1mm] {$\frac{1}{4}$} (b2);
\draw  (b1) edge node[above right, yshift = -1mm, xshift = -1mm] {$\frac{1}{4}$} (b3);
\draw  (b2) edge node[below] {$\frac{1}{2}$} (b3);

\draw  (c1) edge node[above left, yshift = -1mm, xshift = 1mm] {$\frac{1}{6}$} (c2);
\draw  (c1) edge node[above right, yshift = -1mm, xshift = -1mm] {$\frac{1}{6}$} (c3);
\draw  (c2) edge node[below] {$\frac{1}{3}$} (c3);

\draw  (d1) edge node[above left, yshift = -1mm, xshift = 1mm] {$\frac{1}{8}$} (d2);
\draw  (d1) edge node[above right, yshift = -1mm, xshift = -1mm] {$\frac{1}{8}$} (d3);
\draw  (d2) edge node[below] {$\frac{1}{4}$} (d3);
\end{scope}

\begin{scope}[every node/.style={circle, inner sep=2pt}, minimum size = 0.1em]
\node[draw] (e1) at (18, 4.5) {\small $v_{1}$};
\node[draw] (e2) at (16, 3) {\small $v_{2}$};
\node[draw] (e3) at (20, 3) {\small $v_{3}$};
\node (e5) at (18, 1.75) {\small Error: 6e-4};

\end{scope}
\begin{scope}[every edge/.style={draw=black}]

\draw  (e1) edge node[above left, yshift = -1mm, xshift = 1mm] {\small 0.208} (e2);
\draw  (e1) edge node[above right, yshift = -1mm, xshift = -1mm] {\small 0.208} (e3);
\draw  (e2) edge node[below, yshift = -2mm] {\small 0.416} (e3);

\end{scope}
\begin{scope}[every node/.style={circle, inner sep=2pt}, minimum size = 0.1em]
\node[draw] (g1) at (18, -3.25) {\small $v_{1}$};
\node[draw] (g2) at (16, -4.75) {\small $v_{2}$};
\node[draw] (g3) at (20, -4.75) {\small $v_{3}$};
\node (g5) at (18, -6) {\small Error: 0.4e-4};

\end{scope}
\begin{scope}[every edge/.style={draw=black}]

\draw  (g1) edge node[above left, yshift = -1mm, xshift = 1mm] {\small 0.202} (g2);
\draw  (g1) edge node[above right, yshift = -1mm, xshift = -1mm] {\small 0.202} (g3);
\draw  (g2) edge node[below, yshift = -2mm] {\small 0.404} (g3);
\end{scope}

\begin{scope}[every node/.style={circle, inner sep=2pt}, minimum size = 0.1em]
\node[draw] (h1) at (18, 0.625) {\small $v_{1}$};
\node[draw] (h2) at (16, -0.875) {\small $v_{2}$};
\node[draw] (h3) at (20, -0.875) {\small $v_{3}$};
\node (h5) at (18, -2.125) {\small Ground Truth $\{X=5\}$};

\end{scope}
\begin{scope}[every edge/.style={draw=black}]

\draw  (h1) edge node[above left, yshift = -1mm, xshift = 1mm] {\small 0.2} (h2);
\draw  (h1) edge node[above right, yshift = -1mm, xshift = -1mm] {\small 0.2} (h3);
\draw  (h2) edge node[below, yshift = -2mm] {\small 0.4} (h3);
\end{scope}

\begin{scope}[every node/.style={sloped,anchor=south,auto=false}]
    \draw[->](10.25, -0.75) edge node[rotate = 20] {Frobenius} (14.75, 3);
    \draw[->](10.25, -0.75) edge node[below, rotate = -20] {Wasserstein} (14.75, -4.5);
\end{scope}
\end{tikzpicture}
\vspace{-0.6cm}
\caption{We produce experiment in Figure~\ref{fig:1} using the local network regression models instead. The performance is enhanced for both the Frobenius and Wasserstein, however, the Wasserstein remains superior with an error that is $\nicefrac{1}{15}$ of the Frobenius regressors's. }
\label{fig:6}
\vspace{-0.4cm}
\end{figure}

Continued from Section 2, we have the definition of the local Fr\'echet regression model as
\begin{align*}
    m_{L}(x) {=} \argmin_{w \in \mathcal{G}}  \mathbb{E}[s_{L}(X, x)d^{2}(G, w)]
\end{align*}
with its corresponding empirical version
\[
    \hat{m}_{L}(x) := \argmin_{w \in \mathcal{G}} n^{-1}\sum_{i = 1}^{n}s_{iL}(X_{i}, x)d^{2}(G_{i}, w),
\]
where $s_{iL}(x,h) = \frac{1}{\hat{\mu}_{0} - \hat{\mu}_{1}^{T}\hat{\mu}_{2}^{-1}\hat{\mu}_{1}}K_{h}(X_{i} - x)[1 - \hat{\mu}_{1}^{T}\hat{\mu}_{2}^{-1}(X_{i} - x)]$, with $\hat{\mu}_{j} = n^{-1}\sum_{i = 1}^{n}K_{h}(X_{i} - x)(X_{i} - x)^{j}$ for $j = 0, 1, 2$. When applying this model to our toy example from Figure~\ref{fig:1}, we get the results summarized in Figure~\ref{fig:6}

\subsection{Predictions for Large and Random Networks}\label{app:B}

\begin{figure}[h!]
\begin{center}
\begin{tikzpicture}[yscale = 0.5, xscale = 0.5]
\begin{axis}[
    title={Global, Distance from Cycle Graph},
    xlabel={Number of Nodes},
    ylabel={Error [Frobenius Distance]},
    xmin=0, xmax=520,
    ymin=0, ymax=1.2,
    xtick={0, 50, 100, 150, 200, 250, 300, 350, 400, 450, 500},
    ytick={0, 0.1, 0.2, 0.3, 0.4, 0.5, 0.6, 0.7, 0.8, 0.9, 1.0, 1.1},
    legend pos=south east,
    ymajorgrids=true,
    xmajorgrids=true,
    grid style=dashed,
]
\addplot[
    color=blue,
    mark=o,
    ]
    coordinates {
    (20,0.2092547685817618)(40,0.2921609570002253)(60,0.35627025053514355)(80,0.41048624585659754)(100,0.4583333117486596)(120,0.5016372108202715)(140,0.541489019187494)(160,0.5786024753850711)(180,0.6134747681387787)(200,0.6464686827401138)(220,0.6778585578017718)(240,0.7078578183548243)(260,0.7366363808739181)(280,0.7643321426240197)(300,0.7910588415974618)(320,0.8169115971086482)(340,0.8419709164284848)(360,0.8663056566432742)(380,0.8899752565960584)(400,0.913031447418279)(420,0.9355195832239477)(440,0.9574796902373826)(460,0.9789473039127927)(480,0.9999541441781317)(500,1.0205286654864003)
    };

\addplot[
    color=teal,
    mark=o,
    ]
    coordinates {
    (20,1.0739902082375103)(40,1.4995022540143648)
    };

\addplot[
    color=violet,
    mark=o,
    ]
    coordinates {
    (20,0.20927874597799934)(40,0.29219440833007965)(60,0.3563110217084062)(80,0.41053320938674565)(100,0.4583857407253449)(120,0.5016945864631797)(140,0.5415509465385534)(160,0.5786686428323619)(180,0.6135449192953668)(200,0.6465426031585807)(220,0.6779360647824589)(240,0.7079387528736035)(260,0.7367206036466266)(280,0.7644195300622068)(300,0.7911492827592421)(320,0.8170049924195756)(340,0.8420671752957567)(360,0.8664046969303764)(380,0.8900770008535821)(400,0.9131358257333287)(420,0.9356265311033135)(440,0.9575891486835315)(460,0.9790592146880379)(480,1.0000684559200312)
    };
    \legend{Wasserstein, Frobenius, Entropic}  

\node[] at (100, 114.153454967390051) {(20, 1.0739)};
\node[] at (420, 105.041478757080223) {(500, 1.0205)};
\end{axis}
\end{tikzpicture}
\begin{tikzpicture}[yscale = 0.5, xscale = 0.5]
\begin{axis}[
    title={Local, Distance from Cycle Graph},
    xlabel={Number of Nodes},
    ylabel={},
    xmin=0, xmax=520,
    ymin=0, ymax=1.2,
    xtick={0, 50, 100, 150, 200, 250, 300, 350, 400, 450, 500},
    ytick={0, 0.1, 0.2, 0.3, 0.4, 0.5, 0.6, 0.7, 0.8, 0.9, 1.0, 1.1},
    legend pos=north east,
    ymajorgrids=true,
    xmajorgrids=true,
    grid style=dashed,
]
\addplot[
    color=blue,
    mark=o,
    ]
    coordinates {
    (20,0.03628212276761326)(40,0.05065700429016684)(60,0.06177274265231076)(80,0.07117310858688912)(100,0.07946918293929293)(120,0.08697752978886517)(140,0.09388732789521)(160,0.10032233046757782)(180,0.10636874372473343)(200,0.11208947003485219)(220,0.1175320762339096)(240,0.12273356751500468)(260,0.12772340523982237)(280,0.13252549904534705)(300,0.13715956965655715)(320,0.14164210955609371)(340,0.14598707768369976)(360,0.15020641298443826)(380,0.15431042140046103)(400,0.15830807245369236)(420,0.1622072299718088)(440,0.1660148339948379)(460,0.1697370459193648)(480,0.1733793655789667)(500,0.1769467266144332)
    };

\addplot[
    color=teal,
    mark=o,
    ]
    coordinates {
    (20,0.14813669905572516)(40,0.20682795290731762)(60,0.25221255640429086)(80,0.2905934413070806)(100,0.324465618151089)(120,0.3551232308203677)(140,0.383333658523852)(160,0.40960723042547836)(180,0.4342942310055221)(200,0.4576514844951803)(220,0.47987320671001393)(240,0.5011104888498609)(260,0.5214836230365352)(280,0.5410902455159026)(300,0.5600100586297422)(320,0.5783118602345961)(340,0.5960519797912041)(360,0.6132791201647196)(380,0.6300354248716007)(400,0.6463582275449763)(420,0.6622773824055598)(440,0.677823462837093)(460,0.6930209751627813)(480,0.7078927402988323)(500,0.7224580711458324)
    };

\addplot[
    color=violet,
    mark=o,
    ]
    coordinates {
    (20,0.03630395612237378)(40,0.05068745581558306)(60,0.06180985871895423)(80,0.07121585754905116)(100,0.07951690630223876)(120,0.0870297555419678)(140,0.093943696762169)(160,0.1003825587629061)(180,0.10643259811636005)(200,0.11215675571334384)(220,0.11760262619590224)(240,0.12280723761199112)(260,0.12780006814050518)(280,0.13260504220681393)(300,0.1372418928276636)(320,0.14172712168714588)(340,0.14607469609258808)(360,0.1502965624802891)(380,0.15440303281229759)(400,0.15840308140063578)(420,0.16230457769946402)(440,0.16611446824552634)(460,0.16983890991903608)(480,0.17348341650434781)
    };
    \legend{Wasserstein, Frobenius, Entropic}  

\node[] at (420, 74.153454967390051) {(500, 0.7224)};
\node[] at (420, 22.041478757080223) {(500, 0.1769)};
\end{axis}
\end{tikzpicture}
\begin{tikzpicture}[yscale = 0.5, xscale = 0.5]
\begin{axis}[
    title={Global, Distance from Complete Graph},
    xlabel={Number of Nodes},
    ylabel={Error[Frobenius Distance]},
    xmin=0, xmax=190,
    ymin=0, ymax=55,
    xtick={0, 20, 40, 60, 80, 100, 120, 140, 160, 180},
    ytick={0, 5, 10, 15, 20, 25, 30, 35, 40, 45, 50},
    legend pos=south east,
    ymajorgrids=true,
    xmajorgrids=true,
    grid style=dashed,
]
\addplot[
    color=blue,
    mark=o,
    ]
    coordinates {
    (10,0.6295276921001659)(20,1.9456833830199003)(30,3.673387192388302)(40,5.731161431990495)(50,8.072709262673017)(60,10.66708400806998)(70,13.491700703616349)(80,16.529131619401248)(90,19.765397319850894)(100,23.18895949726767)(110,26.790083304910496)(120,30.560411176812327)(130,34.49266564547998)(140,38.58043489800526)(150,42.818013608639966)(160,47.20028197026434)(170,51.722611885873754)
    };

\addplot[
    color=teal,
    mark=o,
    ]
    coordinates {
    (10,3.2310209731289037)(20,9.98612753616591)(30,18.853485255494302)(40,29.414913781804486)(50,41.432796865333515)(60,54.74830194912298)
    };

\addplot[
    color=violet,
    mark=o,
    ]
    coordinates {
    (10,0.6297429899683856)(20,1.9470905440771156)(30,3.6774433165995237)(40,5.739673170464373)(50,8.0877742850019)(60,10.691054954751243)(70,13.527160015461048)(80,16.578872652538035)(90,19.83240949075627)(100,23.276416197747874)(110,26.90133181898723)(120,30.698964100609942)(130,34.66219346986839)(140,38.7847595128109)(150,43.06110255205847)(160,47.48624329095461)(170,52.055689516221655)
    };
    \legend{Wasserstein, Frobenius, Entropic}  

\node[] at (85,520.74830194912298) {(60, 54.7483)};
\node[] at (140,520.7226) {(170, 51.7226)};
\end{axis}
\end{tikzpicture}
\begin{tikzpicture}[yscale = 0.5, xscale = 0.5]
\begin{axis}[
    title={Local, Distance from Complete Graph},
    xlabel={Number of Nodes},
    ylabel={},
    xmin=0, xmax=190,
    ymin=0, ymax=55,
    xtick={0, 20, 40, 60, 80, 100, 120, 140, 160, 180},
    ytick={0, 5, 10, 15, 20, 25, 30, 35, 40, 45, 50},
    legend pos=north east,
    ymajorgrids=true,
    xmajorgrids=true,
    grid style=dashed,
]
\addplot[
    color=blue,
    mark=o,
    ]
    coordinates {
    (10,0.1091521171306692)(20,0.3373568203395382)(30,0.6369187473742669)(40,0.9937107005292843)(50,1.3997053951059175)(60,1.8495370699375444)(70,2.3392897786280487)(80,2.8659417738567683)(90,3.42706919880458)(100,4.020671457262416)(110,4.645060650284207)(120,5.29878768192813)(130,5.980590731637042)(140,6.689358072381335)(150,7.424100473038843)(160,8.183930223989105)(170,8.968044871925885)
    };

\addplot[
    color=teal,
    mark=o,
    ]
    coordinates {
    (10,0.44565806578096673)(20,1.3773969306859049)(30,2.6004807282336224)(40,4.057229487738472)(50,5.714868522102751)(60,7.5514900207160744)(70,9.55110528222628)(80,11.701376965142202)(90,13.992408672717193)(100,16.416032146045218)(110,18.965355847094486)(120,21.63446324962978)(130,24.418202455548165)(140,27.312034384960754)(150,30.31192009785588)(160,33.414235376718764)(170,36.61570338106593)
    };

\addplot[
    color=violet,
    mark=o,
    ]
    coordinates {
    (10,0.10934808288808771)(20,0.3386376165321799)(30,0.6406106274159391)(40,1.0014580828876951)(50,1.4134175847957602)(60,1.871355446606751)(70,2.3715648852109643)(80,2.9112161265794954)(90,3.48806376544963)(100,4.1002746567355635)(110,4.746319174380954)(120,5.424898685194996)(130,6.134895079604542)(140,6.8753344286358455)(150,7.6453600646164865)(160,8.444212164473559)(170,9.27121194978407)
    };
    \legend{Wasserstein, Frobenius, Entropic}  

\node[] at (140, 380) {(170, 36.6157)};
\node[] at (165, 120) {(170, 8.9680)};
\end{axis}
\end{tikzpicture}
\end{center}
\caption{Error between the predicted and ground true graphs for cycle graphs above and complete graphs below. Graphs are shown until the error is greater than $1$ for cycle and $50$ for complete graphs, which occurs around $500$ and $180$ nodes, respectively.}
\label{fig:7}
\end{figure}
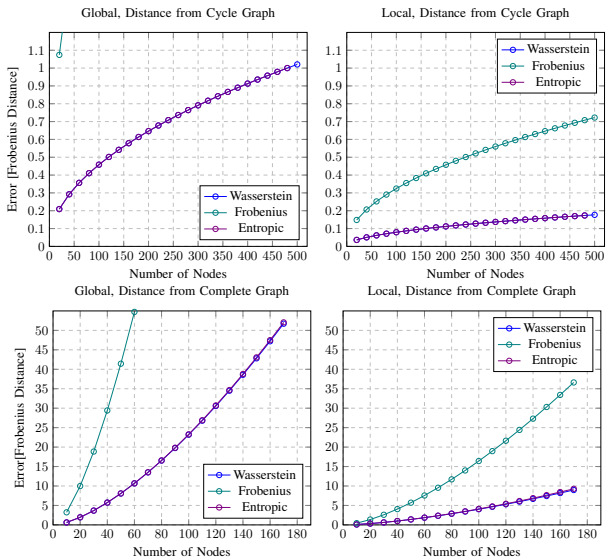

We want to verify that the Wasserstein distance is effective over graphs with a large number of nodes, as the Frobenius distance struggles with graph swelling~\cite{graphMeasures}. Thus, we work over graphs of set structure, specifically the path, cycle, star, and complete. For each, we use a sample set of four graphs with covariates $2$, $4$, $6$, and $8$, with their value denoting edge weights similarly to Figure~\ref{fig:1}. We then output the graph with covariate five and find the error from this graph to the true graph, iterating until said error for the Wasserstein model reaches a certain threshold. As can be seen in Figure~\ref{fig:7}, 
our models use the Frobenius, Wasserstein, and Entropic-Wasserstein distances and have two important findings: the Wasserstein error grows slower than the Frobenius, and the Wasserstein and Entropic-Wasserstein have nearly identical results when using a sufficiently small epsilon ($\varepsilon = 1e^{-5}$).

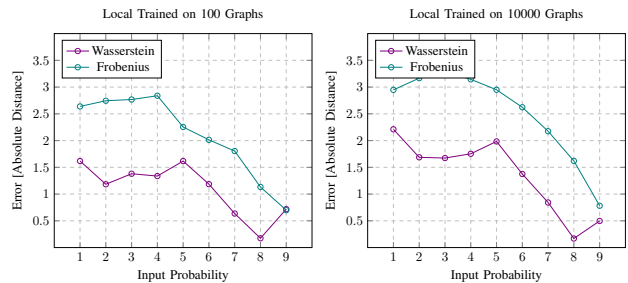
\begin{figure}[h!]

\begin{center}
\begin{tikzpicture}[yscale = 0.5, xscale = 0.5]
\begin{axis}[
    title={Local Trained on 100 Graphs},
    xlabel={Input Probability},
    ylabel={Error [Absolute Distance]},
    xmin=0, xmax=10,
    ymin=0, ymax=4,
    xtick={1, 2, 3, 4, 5, 6, 7, 8, 9},
    ytick={0.5, 1, 1.5, 2, 2.5, 3, 3.5},
    legend pos=north west,
    ymajorgrids=true,
    xmajorgrids=true,
    grid style=dashed,
]
\addplot[
    color=violet,
    mark=o,
    ]
    coordinates {
    (1,1.6182304231830473)(2,1.1834154509278307)(3,1.3801418034109716)(4,1.3366150486905024)(5,1.6172516090377016)(6,1.1854793312949)(7,0.6361838507702782)(8,0.17391233091837321)(9,0.7178434134877953)
    };
\addplot[
    color=teal,
    mark=o,
    ]
    coordinates {
    (1,2.6384581788653936)(2,2.744531440321137)(3,2.7677341926699146)(4,2.8382421330989533)(5,2.253952734574314)(6,2.014090021501808)(7,1.8041225918870918)(8,1.1324283690606656)(9,0.6984971913571076)
    };
    \legend{Wasserstein, Frobenius}  
\end{axis}
\end{tikzpicture}
\begin{tikzpicture}[yscale = 0.5, xscale = 0.5]
\begin{axis}[
    title={Local Trained on 10000 Graphs},
    xlabel={Input Probability},
    ylabel={Error [Absolute Distance]},
    xmin=0, xmax=10,
    ymin=0, ymax=4,
    xtick={1, 2, 3, 4, 5, 6, 7, 8, 9},
    ytick={0.5, 1, 1.5, 2, 2.5, 3, 3.5},
    legend pos=north west,
    ymajorgrids=true,
    xmajorgrids=true,
    grid style=dashed,
]
\addplot[
    color=violet,
    mark=o,
    ]
    coordinates {
    (1,2.2110274855365124)(2,1.687087903114477)(3,1.6737801655640672)(4,1.7530168973494469)(5,1.9843432073275027)(6,1.3755611449487377)(7,0.842337779651058)(8,0.16984358864565685)(9,0.4975057181683713)
    };
\addplot[
    color=teal,
    mark=o,
    ]
    coordinates {
    (1,2.948131630716017)(2,3.1672322334777165)(3,3.286687545731829)(4,3.1450496803967436)(5,2.949915397449553)(6,2.62268630655781)(7,2.17513785299656)(8,1.61935874027826)(9,0.7817509823413697)
    };
    \legend{Wasserstein, Frobenius}  

\end{axis}
\end{tikzpicture}
\end{center}

\caption{Graphs of the distance between the local predicted graphs and the true graph for Fiedler values from $1$ to $9$ for a graph with $10$ nodes.}
\label{fig:8}

\end{figure}

The Wasserstein Distance's improved predictions should also apply to less-deterministic graph structures. This randomness is simulated using Erd\"os-Renyi processes, where networks are generated with a set number of nodes and probability of edge existence~\cite{erdosRenyi}. Graphs of $10$ nodes are generated, each with random probabilities, which almost surely create connected graphs, ranging from $\frac{ln(|V|)}{|V|}$ to $1$. Then, each graph's Fiedler value~\cite{andreotti2018multiplicity}, $\lambda_{F}$, is computed and correlated with each network's Laplacian.  Graphs for connectivity ranging from $1$ to $9$ are output, $\lambda_{F}$ is recomputed, the absolute difference between the expected $\lambda_{F}$, the network's covariate, and its true $\lambda_{F}$ is found. As shown in Figure~\ref{fig:8}, training over $100$ and $10000$ randomly generated graphs, Wasserstein performs significantly better, especially in graphs with Fielder values ranging from $1$ to $5$. This indicates that even in cases where our output networks are randomly generated, details about the graph structure are discoverable more accurately with the Wasserstein distance than the Frobenius.

\subsection{Direct Metric Comparison for Varied Topologies}\label{app:c}

In this section, we separate Figure~\ref{fig:3} into $5$ different figures, comparing performance for each named graph.

\begin{figure}[h!]

\begin{center}
\begin{tikzpicture}[yscale = 0.5, xscale = 0.5]
\pgfplotsset{
width= 430,
height= 225
}
\begin{axis}[
    title={},
    xlabel={},
    ylabel={Error[Frobenius and Wasserstein]},
    xmin=1, xmax=5,
    ymin=0, ymax=10,
    xtick={},
    ytick={0, 1, 2, 3, 4, 5, 6, 7, 8, 9},
    yscale = 0.7,
    legend pos=north west,
    ymajorgrids=true,
    xmajorgrids=true,
    grid style=dashed,
]
\addplot[
    color=green,
    mark=o,
    ]
    coordinates {
    (1.0,0.00022010949935678519)(1.1,0.2871773710780151)(1.2,0.4110387696702948)(1.3,0.5968469430447482)(1.4,0.8752397858816481)(1.5,1.2377398911960826)(1.6,1.6478414943400623)(1.7,2.08804898525544)(1.8,2.549730003708205)(1.9,3.028317105393471)(2.0,3.5213633782156872)(2.1,3.807279135139417)(2.2,4.1236022361426246)(2.3,4.454028358818837)(2.4,4.792131254425465)(2.5,5.13485498399407)(2.6,5.480550902221865)(2.7,5.8282375085062235)(2.8,6.177270382967506)(2.9,6.5271671268404985)(3.0,6.8774995539944)(3.1,7.108923540322379)(3.2,7.443329820456664)(3.3,7.846156157313787)(3.4,8.30154508960987)(3.5,8.799828605893664)(3.6,9.33380794872134)(3.7,9.897111197302385)(3.8,10.482810988285705)(3.9,11.081016991007617)(4.0,11.672189191857468)(4.1,12.4878940592422)(4.2,13.391548163418843)(4.3,14.373006693111021)(4.4,15.425458282178742)(4.5,16.546250331386734)(4.6,17.73866172805889)(4.7,19.01640301594599)(4.8,20.416584533691545)(4.9,22.04607121111242)(5.0,24.33105013454909)
    };
\addplot[
    color=violet,
    mark=o,
    ]
    coordinates {
    (1.0,1.0412364304102084e-09)(1.1,0.019976420285885865)(1.2,0.01845540937386403)(1.3,0.016417754248948313)(1.4,0.03382563770539093)(1.5,0.07876615817202293)(1.6,0.15368973148660814)(1.7,0.25886044412646925)(1.8,0.3940846114779504)(1.9,0.5596765720563823)(2.0,0.7571420474758135)(2.1,0.7991066188513649)(2.2,0.8720065306946054)(2.3,0.9682745025532213)(2.4,1.0879224179188256)(2.5,1.2346756843838662)(2.6,1.4154920323162088)(2.7,1.641600051235642)(2.8,1.931514943326171)(2.9,2.3191842875844486)(3.0,2.881830648422529)(3.1,2.9778752737667773)(3.2,3.130915676741232)(3.3,3.3327598572982637)(3.4,3.585406393604245)(3.5,3.8958593157692505)(3.6,4.27540298123197)(3.7,4.740860369362842)(3.8,5.318483664707195)(3.9,6.054926337025748)(4.0,7.056140837858386)(4.1,7.419200271867119)(4.2,7.986866346037495)(4.3,8.827511889482764)(4.4,10.033979958696527)(4.5,11.731982703339924)(4.6,14.08911060786324)(4.7,17.318304399210383)(4.8,21.652961713003407)(4.9,27.22439607852101)(5.0,33.965876984808304)
    };
    \legend{Frobenius, Wasserstein}
\addplot [color = darkgray, mark=none, style = dashed, opacity = 1] coordinates {(1, 0) (1, 10)};
\addplot [color = darkgray, mark=none, style = dashed, opacity = 1] coordinates {(1.5, 0) (1.5, 10)};
\addplot [color = darkgray, mark=none, style = dashed, opacity = 1] coordinates {(2, 0) (2, 10)};
\addplot [color = darkgray, mark=none, style = dashed, opacity = 1] coordinates {(2.5, 0) (2.5, 10)};
\addplot [color = darkgray, mark=none, style = dashed, opacity = 1] coordinates {(3, 0) (3, 10)};
\addplot [color = darkgray, mark=none, style = dashed, opacity = 1] coordinates {(3.5, 0) (3.5, 10)};
\addplot [color = darkgray, mark=none, style = dashed, opacity = 1] coordinates {(4, 0) (4, 10)};
\addplot [color = darkgray, mark=none, style = dashed, opacity = 1] coordinates {(4.5, 0) (4.5, 10)};
\addplot [color = darkgray, mark=none, style = dashed, opacity = 1] coordinates {(5, 0) (5, 10)};
\end{axis}
\end{tikzpicture}
\end{center}

\caption{An excerpt of Figure~\ref{fig:3} but only considering the Path Graph, comparing the performance of the Frobenius and Wasserstein predictions}
\label{fig:9}

\end{figure}
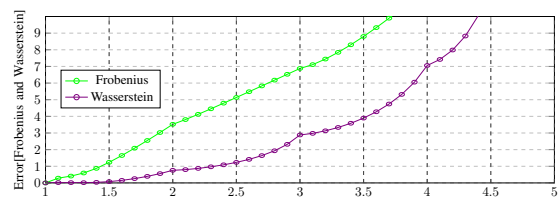

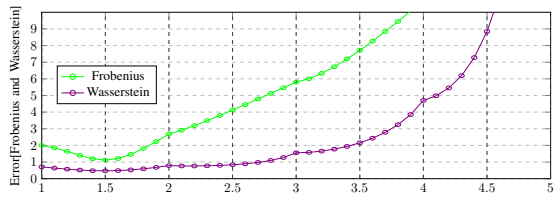
\begin{figure}[ht!]

\begin{center}
\begin{tikzpicture}[yscale = 0.5, xscale = 0.5]
\pgfplotsset{
width= 430,
height= 225
}
\begin{axis}[
    title={},
    xlabel={},
    ylabel={Error[Frobenius and Wasserstein]},
    xmin=1, xmax=5,
    ymin=0, ymax=10,
    xtick={},
    ytick={0, 1, 2, 3, 4, 5, 6, 7, 8, 9},
    yscale = 0.7,
    legend pos=north west,
    ymajorgrids=true,
    xmajorgrids=true,
    grid style=dashed,
]
\addplot[
    color=green,
    mark=o,
    ]
    coordinates {
    (1.0,1.99993424529754)(1.1,1.85542201705415)(1.2,1.6398950242536008)(1.3,1.4036791743625945)(1.4,1.2009622635140016)(1.5,1.1189280754723234)(1.6,1.2111066342271992)(1.7,1.4606876265133855)(1.8,1.8137546414855554)(1.9,2.2288766675621305)(2.0,2.6832815806992025)(2.1,2.9128912930960444)(2.2,3.1880040277415542)(2.3,3.4870315021287133)(2.4,3.800513397806286)(2.5,4.123481458986708)(2.6,4.45300516108422)(2.7,4.787193168730336)(2.8,5.12470247598331)(2.9,5.464426616743357)(3.0,5.8051701193816285)(3.1,6.0039807421590385)(3.2,6.327156814172096)(3.3,6.731572157653174)(3.4,7.1966155451705145)(3.5,7.709538474712115)(3.6,8.261013642001554)(3.7,8.843127997649772)(3.8,9.447728387147032)(3.9,10.063694485751267)(4.0,10.669582959452105)(4.1,11.52136586740005)(4.2,12.462372479898429)(4.3,13.48075075634783)(4.4,14.568718495232865)(4.5,15.723180340785175)(4.6,16.94734938726652)(4.7,18.255184512055514)(4.8,19.684448929671753)(4.9,21.343605503424886)(5.0,23.664319145749523)
    };
\addplot[
    color=violet,
    mark=o,
    ]
    coordinates {
    (1.0,0.7106975598206944)(1.1,0.6368133789322812)(1.2,0.5725304273289211)(1.3,0.5162058085474044)(1.4,0.48080090564148037)(1.5,0.4711721829609772)(1.6,0.4877474271148756)(1.7,0.52916562305208)(1.8,0.5936640572198328)(1.9,0.6798515599378874)(2.0,0.7872444683079749)(2.1,0.76591562055021)(2.2,0.7643393038654338)(2.3,0.7758132383833072)(2.4,0.7995564978813974)(2.5,0.8375432670882361)(2.6,0.8939943203309397)(2.7,0.9759572917291663)(2.8,1.0951396998120089)(2.9,1.2728440278237088)(3.0,1.556845263838028)(3.1,1.5816266285288734)(3.2,1.6565888972015443)(3.3,1.7746922718265026)(3.4,1.9380507346186988)(3.5,2.153346501748949)(3.6,2.4312209776737035)(3.7,2.787458629694349)(3.8,3.246571904411212)(3.9,3.8518685404337916)(4.0,4.701141505905163)(4.1,4.97967391099089)(4.2,5.453652303311223)(4.3,6.1898990082423)(4.4,7.279389443519648)(4.5,8.84560714525636)(4.6,11.05356188164452)(4.7,14.113465285667644)(4.8,18.256678391587513)(4.9,23.61615582396105)(5.0,30.138942089114607)
    };
    \legend{Frobenius, Wasserstein}
\addplot [color = darkgray, mark=none, style = dashed, opacity = 1] coordinates {(1, 0) (1, 10)};
\addplot [color = darkgray, mark=none, style = dashed, opacity = 1] coordinates {(1.5, 0) (1.5, 10)};
\addplot [color = darkgray, mark=none, style = dashed, opacity = 1] coordinates {(2, 0) (2, 10)};
\addplot [color = darkgray, mark=none, style = dashed, opacity = 1] coordinates {(2.5, 0) (2.5, 10)};
\addplot [color = darkgray, mark=none, style = dashed, opacity = 1] coordinates {(3, 0) (3, 10)};
\addplot [color = darkgray, mark=none, style = dashed, opacity = 1] coordinates {(3.5, 0) (3.5, 10)};
\addplot [color = darkgray, mark=none, style = dashed, opacity = 1] coordinates {(4, 0) (4, 10)};
\addplot [color = darkgray, mark=none, style = dashed, opacity = 1] coordinates {(4.5, 0) (4.5, 10)};
\addplot [color = darkgray, mark=none, style = dashed, opacity = 1] coordinates {(5, 0) (5, 10)};
\end{axis}
\end{tikzpicture}
\end{center}

\caption{An excerpt of Figure~\ref{fig:3} but only considering the Cycle Graph, comparing the performance of the Frobenius and Wasserstein predictions}
\label{fig:10}

\end{figure}

When focusing on the cycle graph in Figure~\ref{fig:10} the results are almost identical to the path in Figure~\ref{fig:9}, except we now see the trough of each line occurring at a predictor of 1.5. When we consider the similarity between the path and cycle graphs, this location makes sense, and reassuringly, the variance between the error at predictors $1-2$ is smaller for the Wasserstein, suggesting a greater adherence to the graph structure.

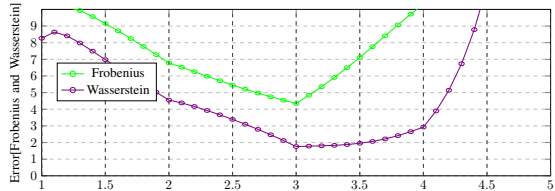
\begin{figure}[h!]
\begin{center}
\begin{tikzpicture}[yscale = 0.5, xscale = 0.5]
\pgfplotsset{
width= 430,
height= 225
}
\begin{axis}[
    title={},
    xlabel={},
    ylabel={Error[Frobenius and Wasserstein]},
    xmin=1, xmax=5,
    ymin=0, ymax=10,
    xtick={},
    ytick={0, 1, 2, 3, 4, 5, 6, 7, 8, 9},
    yscale = 0.7,
    legend pos=north west,
    ymajorgrids=true,
    xmajorgrids=true,
    grid style=dashed,
]
\addplot[
    color=green,
    mark=o,
    ]
    coordinates {
    (1.0,10.197949548766433)(1.1,10.262780256109327)(1.2,10.129407992525856)(1.3,9.931844052505824)(1.4,9.569179852505526)(1.5,9.152704520036087)(1.6,8.71057314594021)(1.7,8.24946417772982)(1.8,7.7731332819347125)(1.9,7.283717886264483)(2.0,6.782329986098317)(2.1,6.52769156889509)(2.2,6.25591181559448)(2.3,5.982045544047632)(2.4,5.712919404349313)(2.5,5.4525564709817615)(2.6,5.2038108096517846)(2.7,4.968848895507687)(2.8,4.749012503446907)(2.9,4.543821424967163)(3.0,4.347413037167639)(3.1,4.8406643400027)(3.2,5.357722143131854)(3.3,5.910305310088594)(3.4,6.497574683217141)(3.5,7.114689148168352)(3.6,7.755199860748969)(3.7,8.41101525194903)(3.8,9.070452261420517)(3.9,9.71246243413743)(4.0,10.287856945880892)(4.1,11.557248996880684)(4.2,12.846201763760904)(4.3,14.156060298748224)(4.4,15.489789595169249)(4.5,16.85284545793573)(4.6,18.2551361127804)(4.7,19.71575187629286)(4.8,21.276148651654676)(4.9,23.046654761239072)(5.0,25.45584413548177)
    };
\addplot[
    color=violet,
    mark=o,
    ]
    coordinates {
    (1.0,8.265645177356436)(1.1,8.638315391470744)(1.2,8.407327061621348)(1.3,7.98019905997171)(1.4,7.4873928125941625)(1.5,6.9785890996614945)(1.6,6.473488921429926)(1.7,5.978905676238455)(1.8,5.495570911703485)(1.9,5.02116948483793)(2.0,4.551814229372045)(2.1,4.3803088419236715)(2.2,4.1655014801241705)(2.3,3.925164813791323)(2.4,3.6657745152131653)(2.5,3.389763057476607)(2.6,3.0978436069532194)(2.7,2.7898880190950024)(2.8,2.46520514141541)(2.9,2.1223011117611748)(3.0,1.7578323777347933)(3.1,1.7828144613672308)(3.2,1.8010441901667562)(3.3,1.8298754498143666)(3.4,1.8789282458468435)(3.5,1.955483694121483)(3.6,2.066081978681666)(3.7,2.2168035075945056)(3.8,2.412478527407302)(3.9,2.6538731874353942)(4.0,2.929868764347944)(4.1,3.8978479340230905)(4.2,5.139240674832116)(4.3,6.732990957105919)(4.4,8.783042711312618)(4.5,11.424991220021056)(4.6,14.830229307656843)(4.7,19.195965469739747)(4.8,24.680392308219233)(4.9,31.142437029207812)(5.0,37.406052314961116)
    };
    \legend{Frobenius, Wasserstein}
\addplot [color = darkgray, mark=none, style = dashed, opacity = 1] coordinates {(1, 0) (1, 10)};
\addplot [color = darkgray, mark=none, style = dashed, opacity = 1] coordinates {(1.5, 0) (1.5, 10)};
\addplot [color = darkgray, mark=none, style = dashed, opacity = 1] coordinates {(2, 0) (2, 10)};
\addplot [color = darkgray, mark=none, style = dashed, opacity = 1] coordinates {(2.5, 0) (2.5, 10)};
\addplot [color = darkgray, mark=none, style = dashed, opacity = 1] coordinates {(3, 0) (3, 10)};
\addplot [color = darkgray, mark=none, style = dashed, opacity = 1] coordinates {(3.5, 0) (3.5, 10)};
\addplot [color = darkgray, mark=none, style = dashed, opacity = 1] coordinates {(4, 0) (4, 10)};
\addplot [color = darkgray, mark=none, style = dashed, opacity = 1] coordinates {(4.5, 0) (4.5, 10)};
\addplot [color = darkgray, mark=none, style = dashed, opacity = 1] coordinates {(5, 0) (5, 10)};
\end{axis}
\end{tikzpicture}
\end{center}

\caption{An excerpt of Figure~\ref{fig:3} but only considering the Star Graph, comparing the performance of the Frobenius and Wasserstein predictions}
\label{fig:11}

\end{figure}

In Figure~\ref{fig:11}, we see the star graph errors for the metrics have the trough at the true predictor for the star graph.

\begin{figure}[h!]
\begin{center}
\begin{tikzpicture}[yscale = 0.5, xscale = 0.5]
\pgfplotsset{
width= 430,
height= 225
}
\begin{axis}[
    title={},
    xlabel={},
    ylabel={Error[Frobenius and Wasserstein]},
    xmin=1, xmax=5,
    ymin=0, ymax=10,
    xtick={},
    ytick={0, 1, 2, 3, 4, 5, 6, 7, 8, 9},
    yscale = 0.7,
    legend pos=north west,
    ymajorgrids=true,
    xmajorgrids=true,
    grid style=dashed,
]
\addplot[
    color=green,
    mark=o,
    ]
    coordinates {
    (1.0,9.695254368845097)(1.1,9.59387816352155)(1.2,9.467811546035671)(1.3,9.337447455484517)(1.4,9.066211462979526)(1.5,8.764232376657684)(1.6,8.458646308901065)(1.7,8.156442422962643)(1.8,7.863006791480501)(1.9,7.58306976688019)(2.0,7.32118560272391)(2.1,6.988433350742246)(2.2,6.669054147242711)(2.3,6.358518525201292)(2.4,6.055664049110697)(2.5,5.760635166152288)(2.6,5.474386592293638)(2.7,5.198704930570173)(2.8,4.936656736465333)(2.9,4.693936544429105)(3.0,4.483302366768181)(3.1,3.780371275118982)(3.2,3.19152422782835)(3.3,2.7212479668865535)(3.4,2.415384640409102)(3.5,2.331734009844371)(3.6,2.490288886043676)(3.7,2.8477685995752458)(3.8,3.3326653144283993)(3.9,3.877437872987302)(4.0,4.409081595129884)(4.1,5.5582412220894915)(4.2,6.776181487495206)(4.3,8.042565335049716)(4.4,9.349936861136857)(4.5,10.698523263906033)(4.6,12.09583266015741)(4.7,13.5606516715036)(4.8,15.136798790843322)(4.9,16.94318553959542)(5.0,19.442222107759605)
    };
\addplot[
    color=violet,
    mark=o,
    ]
    coordinates {
    (1.0,7.17710811804568)(1.1,6.934889908393522)(1.2,6.808125304957969)(1.3,6.718860020056205)(1.4,6.653918014987681)(1.5,6.606386198443339)(1.6,6.570003974902612)(1.7,6.538656563151456)(1.8,6.506426356982182)(1.9,6.467645553696649)(2.0,6.416921557239853)(2.1,6.172407289198745)(2.2,5.933149200345774)(2.3,5.686074255424458)(2.4,5.4254272231519565)(2.5,5.148028692840562)(2.6,4.851896735176737)(2.7,4.535865129615175)(2.8,4.199673752568664)(2.9,3.844595068518153)(3.0,3.4758923755504156)(3.1,3.12379838403011)(3.2,2.780655716109301)(3.3,2.4443748264590752)(3.4,2.1164088124191665)(3.5,1.8001828262428532)(3.6,1.5009495718752248)(3.7,1.2263142053146012)(3.8,0.9876508182297172)(3.9,0.8039395625839774)(4.0,0.7155932967471728)(4.1,0.4720814904163575)(4.2,0.36469909958773883)(4.3,0.4500000389508614)(4.4,0.8065615975489777)(4.5,1.5430615512979386)(4.6,2.8074166106576115)(4.7,4.7919398214639415)(4.8,7.715125714586023)(4.9,11.721573367393503)(5.0,16.835262435567927)
    };
    \legend{Frobenius, Wasserstein}
\addplot [color = darkgray, mark=none, style = dashed, opacity = 1] coordinates {(1, 0) (1, 10)};
\addplot [color = darkgray, mark=none, style = dashed, opacity = 1] coordinates {(1.5, 0) (1.5, 10)};
\addplot [color = darkgray, mark=none, style = dashed, opacity = 1] coordinates {(2, 0) (2, 10)};
\addplot [color = darkgray, mark=none, style = dashed, opacity = 1] coordinates {(2.5, 0) (2.5, 10)};
\addplot [color = darkgray, mark=none, style = dashed, opacity = 1] coordinates {(3, 0) (3, 10)};
\addplot [color = darkgray, mark=none, style = dashed, opacity = 1] coordinates {(3.5, 0) (3.5, 10)};
\addplot [color = darkgray, mark=none, style = dashed, opacity = 1] coordinates {(4, 0) (4, 10)};
\addplot [color = darkgray, mark=none, style = dashed, opacity = 1] coordinates {(4.5, 0) (4.5, 10)};
\addplot [color = darkgray, mark=none, style = dashed, opacity = 1] coordinates {(5, 0) (5, 10)};
\end{axis}
\end{tikzpicture}
\end{center}

\caption{An excerpt of Figure~\ref{fig:3} but only considering the Wheel Graph, comparing the performance of the Frobenius and Wasserstein predictions}
\label{fig:12}
\vspace{-0.5cm}
\end{figure}
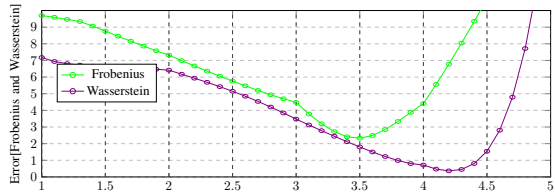

The Wasserstein predictions improved performance for the wheel in Figure~\ref{fig:12}, in particular, because the trough of each line occurs at different inputs for each distance. The Frobenius regressor's trough happens around $0.5$ before the true predictor, and the Wasserstein regressor's trough occurs around $0.2$ after, indicating a smaller in-sample error.

\bibliography{IEEEabrv,source}

\end{document}